\documentclass[sigconf]{acmart}
\AtBeginDocument{%
  }

\usepackage{multirow}
\usepackage{ulem}
\usepackage{bbm}
\usepackage{wrapfig,lipsum}
\usepackage{color}
\usepackage{enumitem}
\usepackage{graphicx}
\usepackage{subfigure}
\setcopyright{acmlicensed}
\copyrightyear{2018}
\acmYear{2018}
\acmDOI{XXXXXXX.XXXXXXX}
\acmConference[Conference acronym 'XX]{Make sure to enter the correct
  conference title from your rights confirmation email}{June 03--05,
  2018}{Woodstock, NY}
\acmISBN{978-1-4503-XXXX-X/2018/06}

\begin{document}

\title{Understanding the Embedding Models on Hyper-relational Knowledge Graph}

\author{Yubo Wang}
\email{ywangnx@connect.ust.hk}
\affiliation{%
  \institution{HKUST}
  \city{Hong Kong}
  \country{China}
}

\author{Shimin Di}
\email{shimin.di@seu.edu.cn}
\affiliation{%
  \institution{Southeast University}
  \city{Nanjing}
  \country{China}
  }

\author{Zhili Wang}
\email{zwangeo@connect.ust.hk}
\affiliation{%
  \institution{HKUST}
\city{Hong Kong}
\country{China}
}

\author{Haoyang Li}
\email{haoyang-comp.li@polyu.edu.hk}
\affiliation{%
 \institution{The Hong Kong Polytechnic University}
 \city{Hong Kong}
 \country{China}
 }

\author{Fei Teng}
\email{fteng@connect.ust.hk}
\affiliation{%
	\institution{HKUST}
	\city{Hong Kong}
	\country{China}
}

\author{Hao Xin}
\email{haoxin@tencent.com}
\affiliation{%
	\institution{Tencent}
	\city{Guangzhou}
	\country{China}
}

\author{Lei Chen}
\email{leichen@cse.ust.hk}
\affiliation{%
	\institution{HKUST \& HKUST(GZ)}
	\city{Guangzhou}
	\country{China}
}

\renewcommand{\shortauthors}{Wang et al.}

\begin{abstract}
	Recently, Hyper-relational Knowledge Graphs (HKGs) have been proposed as an extension of traditional Knowledge Graphs (KGs) to better represent real-world facts with additional qualifiers. As a result, researchers have attempted to adapt classical Knowledge Graph Embedding (KGE) models for HKGs by designing extra qualifier processing modules. However, it remains unclear whether the superior performance of Hyper-relational KGE (HKGE) models arises from their base KGE model or the specially designed extension module. 
Hence, in this paper, we data-wise convert HKGs to KG format using three decomposition methods and then evaluate the performance of several classical KGE models on HKGs. Our results show that some KGE models achieve performance comparable to that of HKGE models. Upon further analysis, we find that the decomposition methods alter the original HKG topology and fail to fully preserve HKG information. 
Moreover, we observe that current HKGE models are either insufficient in capturing the graph's long-range dependency or struggle to integrate main-triple and qualifier information due to the information compression issue. 
To further justify our findings and offer a potential direction for future HKGE research, we propose the FormerGNN framework. 
This framework employs a qualifier integrator to preserve the original HKG topology, and a GNN-based graph encoder to capture the graph's long-range dependencies, followed by an improved approach for integrating main-triple and qualifier information to mitigate compression issues. Our experimental results demonstrate that FormerGNN outperforms existing HKGE models. 
\end{abstract}




\maketitle

\section{Introduction}
\begin{figure*}[t]
	\centering
	\includegraphics[width=1.0\linewidth]{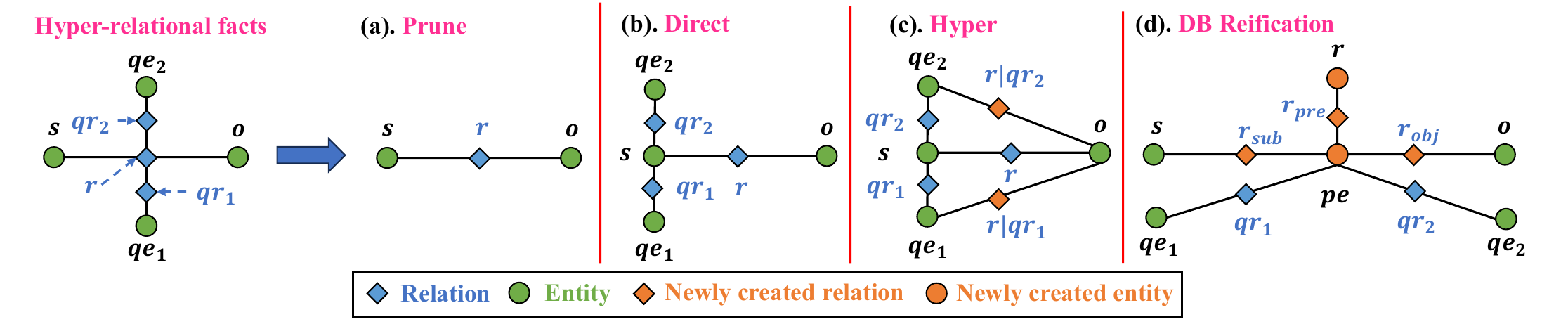}
	\vspace{-18px}
	\caption{
		Illustration of our 3 decomposition methods (sub-figure (a) to (c)) and the DB reification method by \protect\cite{alivanistos2021query} (sub-figure (d)). In (c), $r|qr_1$ and $r|qr_2$ 
		represents new relationships constructed from the main relation $r$ and qualifier relation $qr$. In (d), $r_{pre}, r_{sub}, r_{obj}$ represents the newly generated relationships, and $pe$ represents the newly generated pseudo entity. 
	}
	\label{fig:illustrate_HR}
	\vspace{-10px}
\end{figure*}
\begin{figure}[t]
	\centering
	\includegraphics[width=1.0\linewidth]{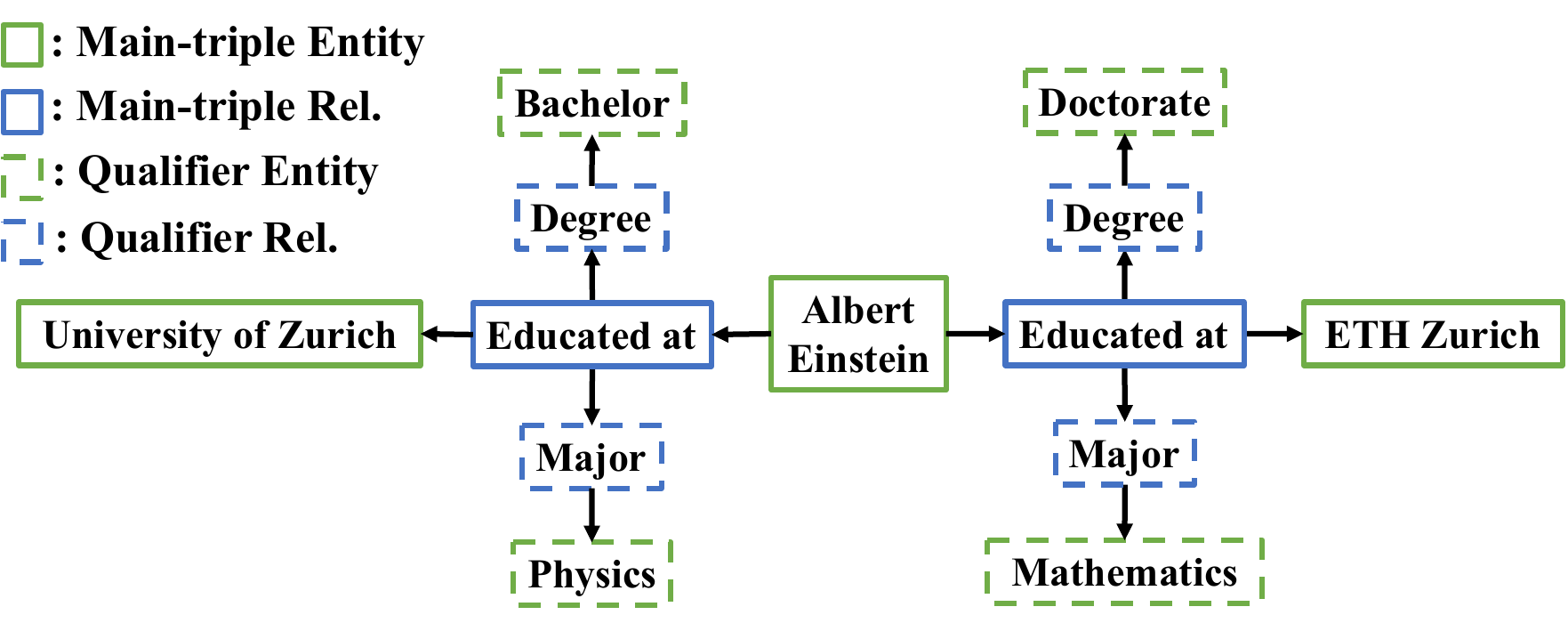}
	\vspace{-15px}
	\caption{
		An HKG example, without the qualifiers with relation \emph{Degree} and \emph{Major}, the difference between entity \emph{University of Zurich} and \emph{ETH Zurich} w.r.t. \emph{Albert Einstein} is hard to detect, as Einstein were educated at both of these universities.
	}
	\label{fig:HKG_example}
	\vspace{-10px}
\end{figure}

\noindent Knowledge Graphs (KGs) \cite{vrandevcic2014wikidata, lehmann2015dbpedia} are powerful tools for organizing and leveraging human knowledge. 
They have facilitated a wide range of web applications, including question answering \cite{yih2015semantic,huang2024rd,zhang2024gail}, 
semantic search \cite{xiong2017explicit,pahuja2024reviving,kamdem2024embedding}, 
and recommendation systems \cite{zhang2016collaborative,zhang2024language,li2024learning,guan2024look}.
Similar to the prune decomposition method shown in \autoref{fig:illustrate_HR}-(a), 
KGs store factual information in the form of triplets $(s,r,o)$, 
where entities $s$ and $o$ are connected by a relation $r$. 
However, this triplet format has been criticized for being insufficient and prone to information loss \cite{wen2016representation,galkin2020message,rosso2020beyond,liu2023self,luo2023hahe,li2024integrating,li2025hyper}. 
To address this limitation, 
researchers have extended KGs by incorporating hyper-relational facts. 
These are represented by adding a qualifier set $\mathcal{Q}$, 
which contains extra entity-relation pairs $(qr,qe)$ to more accurately reflect real-world facts, 
leading to the development of Hyper-relational KGs (HKGs). 
An example of the hyper-relational fact is shown in \autoref{fig:HKG_example}, where the qualifiers with relations \emph{Degree} and \emph{Major} helps to identify the each university that \emph{Albert Einstein} graduated from. 
Given HKGs' practicalities, they were widely applied in many real-world applications. For example, HKGs enable reasoning over time-dependent facts on temporal graphs by incorporating temporal qualifiers into relationships \cite{ding2024temporal}, aiding applications like historical event analysis \cite{hou2023temporal} or forecasting \cite{han2020explainable}. 

Given the importance of HKGs, 
researchers have focused on developing 
embedding models specifically for them, 
known as 
Hyper-relational Knowledge Graph Embedding (HKGE) models. 
Most current HKGE models are extensions of existing KG embedding (KGE) models. 
For instance, 
the well-known GNN-based HKGE model StarE \cite{galkin2020message} is built upon KGE model CompGCN \cite{vashishth2019composition}.

Since most classic HKGE models are extended from KGE models, this inevitably raises a question:
\textit{Are their effectivenesses because of their KGE base model, or the specially designed qualifier processing module?}
Although pioneer work \cite{alivanistos2021query} has found specially designed HKGE models are sometimes not better than KGE with the simple DB reification method. 
It claims that this phenomenon shows the robustness of their specially designed decoder modules, 
while ignoring KGE base models. 
Furthermore, 
it only focuses on one GNN-based KGE model CompGCN \cite{vashishth2019composition} and its HKG extension StarE \cite{galkin2020message}. 
With the development of the KGE and HKGE research, 
various novel models with new technical routines are proposed \cite{zhu2021neural,hu2023hyperformer}, 
do these novel HKGE models also perform similarily with more recent KGE models? 
Therefore,
we dig into the performance of KGE and HKGE models on HKGs to answer this question.

\begin{table*}[h]
	\centering
	\renewcommand{\arraystretch}{1.3}
	\caption{Overview of existing HKG embedding models and how they receive graph topology information and integrate qualifiers. 
		$l$ denotes the number of layers; 
		$\langle \cdot \rangle$ denotes the multi-linear product; 
		$\mathbbm{P}_{r}$ denotes the projection matrix of relation $r$; 
		$HT$ denotes the qualifier integration function of the  HyTransformer \cite{yu2021improving}; Other notations can be refered from \autoref{tab:notation}.
	}
	\vspace{-5px}
	\label{tab:baselines}
	\resizebox{\linewidth}{!}{%
		\begin{tabular}{lllll} 
			\hline\hline
			\multicolumn{1}{l|}{\multirow{2}{*}{\bf Category}}                                                  & \multicolumn{1}{l|}{\multirow{2}{*}{\bf KGE Model}}                           & \multicolumn{1}{l|}{\multirow{2}{*}{\bf HKGE Model}}      & \multicolumn{2}{c}{\bf Key component modeling HKG}                                                                                                                                                                                                                                                                                                                                       \\ 
			\cline{4-5}
			\multicolumn{1}{l|}{}                                                                               & \multicolumn{1}{l|}{}                                                         & \multicolumn{1}{l|}{}                                     & \multicolumn{1}{c|}{\bf Receptive Field}                                                                                                                                           & \multicolumn{1}{c}{\bf Qualifier Integration Function}                                                                                                                                              \\ 
			\hline\hline
			\multicolumn{1}{l|}{\multirow{2}{*}{\begin{tabular}[c]{@{}l@{}}Tensor\\Decomposition\end{tabular}}} & \multicolumn{1}{l|}{SimplE~\cite{kazemi2018simple}}                           & \multicolumn{1}{l|}{HypE~\cite{fatemi2021knowledge}}      & \multicolumn{1}{c|}{\multirow{7}{*}{\begin{tabular}[c]{@{}c@{}}Fact level ($\leq$1-hop): \\$F(s,r,\mathcal{Q})$\end{tabular}}}                                                     & \multicolumn{1}{c}{$\langle h_r,Conv_1(h_s),Conv_2(h_{q_1}),\dots,Conv_{a_r}(h_{q_{a_r-1}}) \rangle$}                                                                                               \\ 
			\cline{2-3}\cline{5-5}
			\multicolumn{1}{l|}{}                                                                               & \multicolumn{1}{l|}{ComplEx~\cite{trouillon2016complex}}                      & \multicolumn{1}{l|}{RAM~\cite{liu2021role}}               & \multicolumn{1}{c|}{}                                                                                                                                                              & \multicolumn{1}{c}{$\Sigma_{i=1}^{a_r}\langle h_r^i,P_r^i[1,:]h_{s},P_r^i[2,:]h_{qe_1},\dots,P_r^i[a_r,:]h_{qe_{a_r-1}} \rangle$}                                                                   \\ 
			\cline{1-3}\cline{5-5}
			\multicolumn{1}{l|}{\multirow{2}{*}{Simple NN}}                                                     & \multicolumn{1}{l|}{\multirow{2}{*}{ConvE~\cite{dettmers2018convolutional}}}  & \multicolumn{1}{l|}{NaLP\_fix~\cite{wang2021link}}        & \multicolumn{1}{c|}{}                                                                                                                                                              & \multicolumn{1}{c}{$MLP(Conv(h_s,h_r,h_{qr_1},h_{qe_1},\cdots))$}                                                                                                                              \\ 
			\cline{3-3}\cline{5-5}
			\multicolumn{1}{l|}{}                                                                               & \multicolumn{1}{l|}{}                                                         & \multicolumn{1}{l|}{Hinge~\cite{rosso2020beyond}}         & \multicolumn{1}{c|}{}                                                                                                                                                              & \multicolumn{1}{c}{$Cat(Conv_1(h_s,h_r),Conv_2(h_s,h_r,h_{qr_1},h_{qe_1},\cdots))$}                                                                                                            \\ 
			\cline{1-3}\cline{5-5}
			\multicolumn{1}{l|}{\multirow{2}{*}{Geometric}}                                                     & \multicolumn{1}{l|}{TransH~\cite{wang2014knowledge}}                          & \multicolumn{1}{l|}{m-TransH~\cite{wang2021link}}         & \multicolumn{1}{c|}{}                                                                                                                                                              & \multicolumn{1}{c}{$\Sigma_{i=1}^{a_r} W_i \mathbbm{P}_{r}(h_i)+b_r$}                                                                                                                               \\ 
			\cline{2-3}\cline{5-5}
			\multicolumn{1}{l|}{}                                                                               & \multicolumn{1}{l|}{-}                                                        & \multicolumn{1}{l|}{ShrinkE~\cite{xiong2023shrinking}}    & \multicolumn{1}{c|}{}                                                                                                                                                              & \multicolumn{1}{c}{$\min_{(qr,qe)\in\mathcal{Q}}(\phi(MLP_1(h_r,h_{qr},h_{qe}),MLP_2(h_r,h_{qr},h_{qe})))$}                                                                                         \\ 
			\cline{1-3}\cline{5-5}
			\multicolumn{1}{l|}{\multirow{3}{*}{Transformer}}                                                   & \multicolumn{1}{l|}{\multirow{3}{*}{-}}                                       & \multicolumn{1}{l|}{HyTransformer~\cite{yu2021improving}} & \multicolumn{1}{c|}{}                                                                                                                                                              & \multicolumn{1}{c}{$Trm(Cat(h_s,h_r,h_{qr_1},h_{qe_1},\cdots))$}                                                                                                                                    \\ 
			\cline{3-5}
			\multicolumn{1}{l|}{}                                                                               & \multicolumn{1}{l|}{}                                                         & \multicolumn{1}{l|}{HyperFormer~\cite{hu2023hyperformer}} & \multicolumn{1}{c|}{\begin{tabular}[c]{@{}c@{}}Node level ($=$1-hop):\\$\Sigma_{r',\mathcal{Q}'\in \mathcal{N}(s)}(s,r',\mathcal{Q}')$\end{tabular}}                               & \multicolumn{1}{c}{$Conv(				HT(s,r,\mathcal{Q}),				\gamma\frac{\Sigma_{(s,r',\mathcal{Q}') \in \mathcal{N}(s)}Trm(F(s,r',\mathcal{Q}'))}{|\mathcal{N}(s)|},\gamma (h_{qr_1}, h_{qe_1},\cdots))$}  \\ 
			\cline{3-5}
			\multicolumn{1}{l|}{}                                                                               & \multicolumn{1}{l|}{}                                                         & \multicolumn{1}{l|}{HAHE~\cite{luo2023hahe}}              & \multicolumn{1}{c|}{\multirow{3}{*}{\begin{tabular}[c]{@{}c@{}}Graph level ($>$1-hop):\\$\big(\Sigma_{r',\mathcal{Q}'\in \mathcal{N}(s)}(s,r',\mathcal{Q}')\big)^l$\end{tabular}}} & \multicolumn{1}{c}{$MLP(Cat(Trm(Cat(h_s,h_r,h_{qr_1},h_{qe_1},\cdots)),h_r,b_e,b_r))$}                                                                                                                                                                      \\ 
			\cline{1-3}\cline{5-5}
			\multicolumn{1}{l|}{\multirow{2}{*}{GNN}}                                                           & \multicolumn{1}{l|}{\multirow{2}{*}{CompGCN~\cite{vashishth2019composition}}} & \multicolumn{1}{l|}{StarE~\cite{galkin2020message}}       & \multicolumn{1}{c|}{}                                                                                                                                                              & \multicolumn{1}{c}{$\phi(h_s,\gamma(h_r,h_{qr_1},h_{qe_1},\cdots))$}                                                                                                                                \\ 
			\cline{3-3}\cline{5-5}
			\multicolumn{1}{l|}{}                                                                               & \multicolumn{1}{l|}{}                                                         & \multicolumn{1}{l|}{QUAD~\cite{shomer2023learning}}       & \multicolumn{1}{c|}{}                                                                                                                                                              & \multicolumn{1}{c}{$\gamma(\phi(h_s,h_r),h_{qr_1},h_{qe_1},\cdots)$}                                                                                                                                \\ 
			\hline\hline
			&                                                                               &                                                           &                                                                                                                                                                                    &                                                                                                                                                                                                     \\
			&                                                                               &                                                           &                                                                                                                                                                                    &                                                                                                                                                                                                     \\
			&                                                                               &                                                           &                                                                                                                                                                                    &                                                                                                                                                                                                     \\
			&                                                                               &                                                           &                                                                                                                                                                                    &                                                                                                                                                                                                     \\
			&                                                                               &                                                           &                                                                                                                                                                                    &                                                                                                                                                                                                     \\
			&                                                                               &                                                           &                                                                                                                                                                                    &                                                                                                                                                                                                     \\
			&                                                                               &                                                           &                                                                                                                                                                                    &                                                                                                                                                                                                     \\
			&                                                                               &                                                           &                                                                                                                                                                                    &                                                                                                                                                                                                     \\
			&                                                                               &                                                           &                                                                                                                                                                                    &                                                                                                                                                                                                     \\
			&                                                                               &                                                           &                                                                                                                                                                                    &                                                                                                                                                                                                     \\
			&                                                                               &                                                           &                                                                                                                                                                                    &                                                                                                                                                                                                     \\
			&                                                                               &                                                           &                                                                                                                                                                                    &                                                                                                                                                                                                     \\
			&                                                                               &                                                           &                                                                                                                                                                                    &                                                                                                                                                                                                     \\
			&                                                                               &                                                           &                                                                                                                                                                                    &                                                                                                                                                                                                     \\
			&                                                                               &                                                           &                                                                                                                                                                                    &                                                                                                                                                                                                    
		\end{tabular}
	}
	\vspace{-195px}
\end{table*}

To make fair comparasions between KGE and HKGE models on HKGs, 
we need to make KGE models runnable on HKGs. 
Previous researchers apply 
\texttt{model-wise}
extension on
KGE model,
which focus on upgrading KGE models to 
model HKG \cite{fatemi2021knowledge, liu2021role, wang2021link, rosso2020beyond, wen2016representation, galkin2020message,shomer2023learning}.
In this paper, 
we turn to another view to
\texttt{data-wise}
extend HKG to KG format. 
As shown in \autoref{fig:illustrate_HR}-(d),
	DB reification \cite{alivanistos2021query}
	converts main-triple relation into entity and then add pseudo nodes and relations to represent hyper-relational facts.
	It cannot preserve the most important main triplets, 
and hinders us from incrementally studying the effect of each HKG component. 
Consequently, we propose three decomposition methods that can convert HKG to KG's triplet format while keeping the original format of HKG main-triples and preserving the HKG information to different extents. 
In this paper, 
we divide the preserved HKG information into 3 parts: 
main triple, intra-qualifier, and inter-qualifier information (The detailed definition of them are in \autoref{sec:decomp}). 
Through such a data-wise way, we can easily test conventional KGE models on HKGs.
As discussed in \autoref{sec:performance-hkg},
our preliminary experimental results demonstrate that 
KGE models can still achieve comparable performance to those specially designed HKGE models on HKGs.

As the innovation of most of the HKGE model is their specially designed qualifier capturing modules, 
two questions naturally arise from the results:
	\begin{enumerate}
		\item \textit{Is it necessary to develop various HKGEs?}
		\item \textit{What information are current HKGEs overlooking?}
	\end{enumerate}
	In \autoref{sec:futher}, 
	we further delve deeper into the model performance comparison between KGE and HKGE models to answer these two questions.
Firstly, we examine how these models capture qualifiers, as the key distinction between KGE and HKGE models lies in the inclusion of specialized modules designed to handle qualifiers. 
Our findings show that HKGE models are still necessary, 
as KGE models with decomposition methods shift the original HKG topology and are inadequate for capturing qualifiers in HKGs. 
As a result, on HKG, KGE models suffer from \textit{information loss}, such as semantic shifts and misinterpretations.
Secondly, 
to identify the information missed by HKGE models, 
we analyze the graph reception approach of both HKGE and KGE models. 
By ``graph reception," 
we refer to how models receive and integrate information from the HKG, 
including both main-triple and qualifier information. 
Our analysis reveals that current HKGE models often suffer from a  \textit{limited receptive field}, 
which restricts their ability to capture the graph's multi-hop topology. 
This limitation weakens their capacity to model the graph's long-range dependency, which is crucial for many tasks. 
Furthermore, 
while some HKGE models have broader receptive fields, they still compress qualifier information into fixed-sized main triple embedding matrices, 
leading to the \textit{information compression issue}.

	The above experiments are all based on an experimental observation of current works. 
	Therefore, we propose a new framework FormerGNN to incorporate several key mechanisms to justify the above speculations on qualifier capture and graph reception.
Firstly, FormerGNN employs a transformer-based qualifier integrator to capture the original topology of the HKG within the 1-hop neighborhood, eliminating the need for decomposition methods. 
Secondly, it uses a GNN-based graph encoder to maintain a broad receptive field for main triples and effectively capture the graph's long-range dependencies. 
Finally, FormerGNN performs predictions jointly with the main-triple and qualifier embedding matrix, 
without the need to compress one into another, mitigating the inofmation compression issue. 
Experimental results demonstrate that FormerGNN achieves state-of-the-art performance across all HKGE models on the Cleaned JF17k and WD50k benchmarks.

Our contributions can be summarized as follows:
\begin{itemize}[leftmargin=*]
	\item Although various HKG embedding models have been developed, 
	the source of their notable performance is yet cleared. In this papers, 
	we conduct a comprehensive analysis, 
	to systematically evaluate whether their performance is due to their specially designed qualifier processing module, 
	or their KGE base model.
	
	
	\item To data-wise extend HKG to KG format and analyse the performance of KGE model on HKG, 
	we proposed three HKG decomposition methods that can preserve HKG information to different extents, 
	while maintaining HKG's original main-triple topology.

	\item To justify our claims while also providing a possible better HKGE research direction, 
	we propose FormerGNN\footnote{Github: https://github.com/Wyb0627/GraphEmbedding}, which leverages a transformer based qualifier integrator that can capture HKG's original topology, a GNN graph encoder to capture the graph's long range dependency, 
	and predict jointly with all obtained embeddings and mitigate GNNs' information compression issue. 
\end{itemize}

\section{Preliminary}

\subsection{Further HKG Backgrounds}
As illustrated, 
KG's triplet format alone 
may suffer from information loss~\cite{wen2016representation}, 
 and are hard to represent the facts with more than two entities~\cite{rosso2020beyond}.
Hence, to address the shortcomings of conventional KGs, the facts within these KGs have been extended to 
hyper-relational facts~\cite{galkin2020message}
(\autoref{fig:illustrate_HR}) by adding multiple key-value pairs named qualifiers, and form HKGs. 
In this paper, 
following \cite{galkin2020message}, 
we denote the hyper-relational facts $F(s, r, o, \mathcal{Q})$ as a main triplet along the qualifier set which contains $n$ qualifiers: $\mathcal{Q}:=\{(qr_i, qe_i)\}_{i=1}^n$. 
For triplet facts, 
they can be considered hyper-relational facts with the number of qualifiers $n=0$, 
in this paper, we represent them as subject-relation-object triples: $(s, r, o)$, 
and formally formulate HKGs as: $\mathcal{G}(\mathcal{V},\mathcal{R},\mathcal{F})$, where $s, o, qe_1, \dots, qe_n \in \mathcal{V}$, $r, qr_1, \dots, qr_n \in \mathcal{R}$, 
and 
$
F(s,r,\mathcal{Q}):=(s,r,o, \{(qr_i, qe_i)\}_{i=1}^n) \in \mathcal{F}
$, 
with $\mathcal{V}$ representing the entity set (node set),   
$\mathcal{R}$ representing the relation set (edge set), 
and $\mathcal{F}$ representing the fact set. 
In this paper, we denote scalars by lowercase letters $r$, and its respective vector by $h_r$. 

\begin{table}[t]
	\centering
	\caption{Summary of Important Notations.}
	\label{tab:notation}
	\vspace{-5px}
	\setlength\tabcolsep{4pt}
	\resizebox{\linewidth}{!}{
		\begin{tabular}{l|p{7cm}}
			\hline\hline
			\textbf{Notations}   & \textbf{Meanings}  \\ 
			\hline
			
			$s$ & The subject entity of main-triple. \\  \hline
			
			$r$ & The relation of main-triple.  \\ 		\hline
			$qe$ & The qualifier entity.   \\ \hline
			$qr$ & The qualifier relation.   \\ \hline
			$h_*$ & The embedding matrix of an entity or a relation.   \\ \hline
			$W$ & The learnable weight matrix. \\ \hline
			$b$ & The learnable bias. \\ \hline
			$\mathcal{G}(\mathcal{V},\mathcal{R},\mathcal{F})$ & The knowledge graph with node set $\mathcal{V}$, edge set $\mathcal{R}$, and fact set $\mathcal{F}$.  \\ \hline
			
			$F(s,r,\mathcal{Q})$ & The hyper-relational fact. \\\hline
			
			$\mathcal{Q}$  &  The qualifier set. \\ \hline
			$\mathcal{N(\cdot)}$ & The 1-hop graph neighborhood of an entity. \\ 		\hline
			$Conv$ &  The convolutional operation. \\ \hline
			$Cat$ &  The matrix concatenation operation. \\ \hline
			$\phi$ & The message passing function. \\ 
			\hline
			$\oplus$ & The aggregration function.\\
			\hline
			$\gamma$ &  The composition operation. \\ \hline
			$Trm$ &   	The transformer layers.\\ \hline
			$[MSK]$ &   	The MASK token of the transformer.\\
			\hline\hline
		\end{tabular}
	}
	\vspace{-10px}
\end{table}
\subsection{Specially Designed HKG Embedding Models}
In this paper, 
we categorize current HKGE methods into five categories based on their approach to integrating qualifiers. 
We will primarily focus on the following HKGE models, which represent the key methodologies in these categories.

\noindent \textbf{Tensor Decomposition Models} 
These models decompose the basis matrix into several matrices, to represent entities and relations in KGs and HKGs. 
For example, RAM~\cite{liu2021role} is a tensor decomposition model based on the classic bilinear KGE model ComplEx~\cite{trouillon2016complex}. Compared with ComplEx, it further considers the main triple and qualifiers as different roles of the hyper-relational facts and models them with role-specific pattern matrices.
On the other hand, HypE~\cite{fatemi2021knowledge} extends another bilinear KGE model SimplE~\cite{kazemi2018simple} by considering additional qualifier positions and applying a abstract relation to loosely represent a combination of all qualifier relations of the original hyper-relational fact. 

This category of models treats each fact independently and aggregates the information at the end.
Consequently, according to \autoref{tab:baselines}, they only directly receive information within facts and have a receptive field less or equal to the graph's 1-hop neighborhood. 

\noindent \textbf{Simple NN-based Models} 
These models apply simple, lightweight neural network structures, such as Convolutional Neural Networks (CNNs) or Multi-layer Perceptrons (MLPs), directly to the feature matrices of facts to learn their semantic information. 
For example, NaLP~\cite{wang2021link} and Hinge~\cite{rosso2020beyond} extend their base KGE model, ConvE~\cite{dettmers2018convolutional}, by further appending the feature matrices of qualifiers to those of main-triples to create the feature matrices of hyper-relational facts. 
Then they apply CNN or MLP layers to these feature matrices to encode hyper-relational facts and then aggregate these facts w.r.t. their subject entities and relations for the graph's entity and relation embeddings.

Similar to Tensor Decomposition Models, Simple NN-based models also treat each fact independently and have receptive field less or equal to the graph's 1-hop neighborhood. 

\noindent \textbf{Geometric Models} 
These models embed KGs or HKGs by projecting entities and relations onto hyperplanes and trying to optimize the model based on the geometric cost functions. 
KGE model TransH~\cite{wang2014knowledge} only applies the cost function for the main triple. 
In order to model HKGs, m-TransH~\cite{wen2016representation} 
extends TransH by considering the separate cost functions for entity-relation pairs within hyper-relational facts and calculating a weighted sum of their cost together. 
ShrinkE \cite{xiong2023shrinking} is a model specifically designed to model HKG, 
it maps main triples to boxes in the geometric hyperplane, 
then considers the qualifiers as geometric shrinkings of the box and uses this extra information further to reduce the search space for the answer.

As Geometric Models model each fact independently, according to \autoref{tab:baselines}, they also have a receptive field less or equal to the graph's 1-hop neighborhood. 

\noindent \textbf{Transformer-based Models} This category of models focuses on applying the dense parameter transformer \cite{vaswani2017attention} on feature hyper-relational facts' feature matrices to integrate qualifiers and the main-triple and directly carry out prediction with the transformer's last layer output. Among this category, HyTransformer~\cite{yu2021improving} is a simple method that treats hyper-relational fact as a sequence of entities and relations, and does not apply any special approach to deal with qualifiers: 
\begin{align}
	h_o=h_{[MSK]}
	&=Trm(F(s,r,[MSK],\mathcal{Q}))\\&
	=Trm(Cat(h_s,h_r,h_{[MSK]},h_{qr_1},h_{qe_1},\cdots)), \notag
\end{align}
\label{eq:hytransformer}
where the embedding generated from the mask token $[MSK]$ is considered as the embedding $h_o$ of the predict target entity $o$, and $Cat$ denotes the concatenation operation. 
HyTransformer substitutes the GNN in StarE with Layer Normalization and Dropout layers and only uses the Transformer decoder for prediction, to achieve the fact-level receptive field. Furthermore, a qualifier prediction subtask is applied to enhance its link prediction performance. 
GRAN~\cite{wang2021link} treats the hyper-relational fact as a heterogeneous graph and maintains the connectivity between qualifiers through the attention mechanism of the transformer.

Building upon HyTransformer, HyperFormer~\cite{hu2023hyperformer} further incorporates additional aggregators to capture local-level sequential information. It also applies convolutional operations to integrate information from multiple aggregators:
\begin{align}
	h_o=&Conv(h_{[MSK]},\gamma_{rq}(h_{qr_1},h_{qe_1},\cdots), \\ &\gamma_{en}(\Sigma_{(s,r',\mathcal{Q}')\in \mathcal{N}(s)}Trm(s,r', \mathcal{Q}')/|\mathcal{N}(s)|)), \notag
\end{align}
where $Conv$ denote the convolutional operator, $h_{[MSK]}$ is the same as in \autoref{eq:hytransformer}, 
$\gamma_{rq}$ and $\gamma_{en}$ are composition operations of HyperFormer's relation qualifier and entity neighbor aggregator, respectively.
Consequently, HyperFormer can directly receive graph information within the graph's 1-hop neighborhood, rather than the fact only. 

On the other hand, to increase the model's graph receptive field, HAHE \cite{luo2023hahe} utilizes the graph transformer \cite{yun2019graph,shehzad2024graph,yuan2025survey} to aggregrate the HKG's global dependency into node embeddings, and then model the HKG's local dependency as bias injected to the hyper-relational fact embedding:  
\begin{align}
	MLP(Cat(Trm(Cat(h_s,h_r,h_{qr_1},h_{qe_1},\cdots)),h_r,b_e,b_r)). 
\end{align}
where $b_e$ and $b_r$ denotes the entity and relation bias, respectively.

\begin{table*}
	\centering
	\caption{Performance of KG/HKG embedding methods on two most representative HKG benchmarks.}
	\label{tab:metrics}
	\vspace{-8px}
	\scalebox{0.9}{
		\resizebox{\linewidth}{!}{%
			\begin{tabular}{l|l|l|c|ccc|ccc} 
				\hline\hline
				\multirow{2}{*}{\bf Category}                                                   & \multirow{2}{*}{\bf KGE Model} & \multirow{2}{*}{\begin{tabular}[c]{@{}l@{}}\bf HKGE Model \\\bf / Modules\end{tabular}} & \multirow{2}{*}{\begin{tabular}[c]{@{}c@{}}\bf Receptive \\\bf Field\end{tabular}} & \multicolumn{3}{c|}{\bf Cleaned JF17k \cite{galkin2020message}}                                      & \multicolumn{3}{c}{\bf WD50K \cite{galkin2020message}}                                                \\ 
				\cline{5-10}
				&                                &                                                                                         &                                                                                   & \multicolumn{1}{c|}{MRR} & \multicolumn{1}{c|}{H@1} & H@10                  & \multicolumn{1}{c|}{MRR} & \multicolumn{1}{c|}{H@1} & H@10                   \\ 
				\hline\hline
				\multirow{2}{*}{\begin{tabular}[c]{@{}l@{}}Tensor\\Decomposition\end{tabular}}  & SimplE~\cite{kazemi2018simple}                         & HypE~\cite{fatemi2021knowledge}                                                                                    & \multirow{8}{*}{$\leq$ 1-hop}                                                     & 0.295                    & 0.211                    & 0.454                 & -                        & -                        & -                      \\
				& ComplEx~\cite{trouillon2016complex}                        & RAM~\cite{liu2021role}                                                                                     &                                                                                   & 0.342                    & 0.257                    & 0.505                 & 0.300                    & 0.236                    & 0.422                  \\ 
				\cline{1-3}\cline{5-10}
				\multirow{2}{*}{Simple NN}                                                      & \multirow{2}{*}{ConvE~\cite{dettmers2018convolutional}}         & NaLP\_fix~\cite{wang2021link}                                                                               &                                                                                   & 0.067                    & 0.044                    & 0.108                 & 0.177                    & 0.131                    & 0.264                  \\
				&                                & Hinge~\cite{rosso2020beyond}                                                                                   &                                                                                   & 0.272                    & 0.178                    & 0.461                 & 0.243                    & 0.176                    & 0.377                  \\ 
				\cline{1-3}\cline{5-10}
				\multirow{2}{*}{Geometric}                                                      & TransH~\cite{wang2014knowledge}                         & m-TransH~\cite{wen2016representation}                                                                                &                                                                                   & 0.262                    & 0.188                    & 0.393                 & -                        & -                        & -                      \\
				& -                              & ShrinkE~\cite{xiong2023shrinking}                                                                                 &                                                                                   & 0.343                    & 0.257                    & 0.510                 & 0.323                    & 0.260                    & 0.441                  \\ 
				\cline{1-3}\cline{5-10}
				\multirow{3}{*}{Transformer}                                                    & \multirow{3}{*}{-}             & HyTransformer~\cite{yu2021improving}                                                                           &                                                                                   & 0.368                    & 0.265                    & 0.592                 & 0.348                    & 0.272                    & 0.489                  \\
				& & HAHE~\cite{luo2023hahe}                                                                           &                                                                                   & 0.405                    & 0.310                    & 0.585                 & 0.360                    & 0.288                    & 0.496                   \\
				&                                & HyperFormer~\cite{hu2023hyperformer}                                                                             &                                                                                   & \bf 0.425                & \bf 0.328                & \bf 0.613             & \bf0.372                 & \bf 0.293                & \bf 0.522              \\ 
				\hline
				\multirow{2}{*}{Neighbor-GNN}                                                   & \multirow{2}{*}{CompGCN~\cite{vashishth2019composition}}       & StarE~\cite{galkin2020message}                                                                                   & \multirow{2}{*}{$>$1-hop}                                                         & 0.381                    & 0.284                    & 0.577                 & 0.346                    & 0.268                    & 0.494                  \\
				&                                & QUAD~\cite{shomer2023learning}                                                                                    &                                                                                   & 0.402                    & 0.305                    & 0.602                 & 0.349                    & 0.275                    & 0.487                  \\ 
				\hline\hline
				\multirow{3}{*}{\begin{tabular}[c]{@{}l@{}}Tensor \\Decomposition\end{tabular}} & \multirow{3}{*}{ComplEx~\cite{trouillon2016complex}}       & Pruned                                                                                  & \multirow{9}{*}{$\leq$1-hop}                                                      & 0.112                    & 0.079                    & 0.173                 & 0.103                    & 0.058                    & 0.194                  \\
				&                                & Direct                                                                                  &                                                                                   & 0.103                    & 0.074                    & 0.156                 & 0.102                    & 0.057                    & 0.188                  \\
				&                                & Hyper                                                                                   &                                                                                   & 0.111                    & 0.076                    & 0.174                 & 0.134                    & 0.087                    & 0.225                  \\ 
				\cline{1-3}\cline{5-10}
				\multirow{3}{*}{Simple NN}                                                      & \multirow{3}{*}{ConvE~\cite{dettmers2018convolutional}}         & Pruned                                                                                  &                                                                                   & 0.120                    & 0.074                    & 0.210                 & 0.042                    & 0.014                    & 0.096                  \\
				&                                & Direct                                                                                  &                                                                                   & 0.131                    & 0.084                    & 0.219                 & 0.033                    & 0.015                    & 0.066                  \\
				&                                & Hyper                                                                                   &                                                                                   & 0.138                    & 0.078                    & 0.248                 & 0.041                    & 0.017                    & 0.103                  \\ 
				\cline{1-3}\cline{5-10}
				\multirow{3}{*}{Geometric}                                                      & \multirow{3}{*}{TransH~\cite{wang2014knowledge}}        & Pruned                                                                                  &                                                                                   & 0.118                    & 0.031                    & 0.307                 & 0.167                    & 0.008                    & 0.347                  \\
				&                                & Direct                                                                                  &                                                                                   & 0.130                    & 0.037                    & 0.335                 & 0.169                    & 0.008                    & 0.352                  \\
				&                                & Hyper                                                                                   &                                                                                   & 0.136                    & 0.042                    & 0.345                 & 0.171                    & 0.008                    & 0.354                  \\ 
				\hline
				\multirow{3}{*}{Path-GNN}                                                       & \multirow{3}{*}{NBFNet~\cite{zhu2021neural}}        & Pruned                                                                                  & \multirow{6}{*}{$>$1-hop}                                                         & \bf \color{blue}0.415    & \bf \color{blue}0.313    & \bf \color{blue}0.630 & \bf \color{blue}0.360    & 0.271                    & \bf \color{blue}0.530  \\
				&                                & Direct                                                                                  &                                                                                   & 0.412                    & 0.308                    & 0.623                 & 0.351                    & 0.261                    & 0.523                  \\
				&                                & Hyper                                                                                   &                                                                                   & 0.396                    & 0.299                    & 0.580                 & 0.346                    & 0.257                    & 0.517                  \\ 
				\cline{1-3}\cline{5-10}
				\multirow{3}{*}{Neighbor-GNN}                                                   & \multirow{3}{*}{CompGCN~\cite{vashishth2019composition}}       & Pruned                                                                                  &                                                                                   & 0.383                    & 0.291                    & 0.577                 & 0.348                    & 0.273                    & 0.487                  \\
				&                                & Direct                                                                                  &                                                                                   & 0.382                    & 0.290                    & 0.579                 & 0.350                    & \bf \color{blue}0.277    & 0.488                  \\
				&                                & Hyper                                                                                   &                                                                                   & 0.389                    & 0.294                    & 0.590                 & 0.350                    & \bf \color{blue}0.277    & 0.488                  \\ 
				\hline\hline
			\end{tabular}
	}}
	\vspace{-5px}
\end{table*}

\noindent \textbf{Graph Neural Network Models} 
Current Graph Neural Network (GNN)-based KGE models can be classified into two groups: neighbor-aggregated models, and path-aggregated models. 
Given a KG
$\mathcal{G}(\mathcal{V},\mathcal{R},\mathcal{F})$,
neighbor-aggregated KGE model (Neighbor-GNN in \autoref{tab:metrics}) CompGCN \cite{vashishth2019composition} extends GNNs
to multi-relational graphs and propose the node representation 
\begin{equation}
	h_o=act(\bigoplus_{(s,r)\in\mathcal{N}(s)}(\{W_{\lambda(r)}\phi(h_s,h_r) \})), 
\end{equation}
where 
$\bigoplus$ denotes the aggregation function; 
$h_o$ and $h_r$ 
denote the embedding vector of entity $o$ and relation $r$ respectively, 
$\phi(\cdot): \mathbb{R}^d \times \mathbb{R}^d \to \mathbb{R}^d$
is the message passing function,
$W_{\lambda(r)}$ 
is the direction-specific weighting matrix,
and 
$\mathcal{N}(s)$ 
denotes the neighborhood of node $s$.

Based on CompGCN, HKGE model StarE~\cite{galkin2020message} was proposed. It adapts to HKGs by further generate embeddings $h_q$ for each qualifier pair $q$ as the embedding aggregation of the qualifier entity $h_{qe}$ and qualifier relation $h_{qr}$: 
\begin{equation}
	h_q=W(sum\{\phi_q(h_{qr},h_{qe})\}_{(qr,qe)\in\mathcal{Q}}).
\end{equation}
Next, StarE would aggregate $h_q$ into the main triple's relation embedding matrix $h_r$: 
\begin{equation}
	h_o=act(\bigoplus_{(s,r)\in\mathcal{N}(s)}(\{W_{\lambda(r)}\phi(h_s,\gamma(h_r, h_q)) \})),
\end{equation}
where $\phi_q$ can be any entity-relation function akin to $\phi$; $\gamma$ can be any function that combines multiple embeddings, such as the weighted sum or simple average.

Based on StarE, 
QUAD~\cite{shomer2023learning}
aggregates the qualifier information into the entity and relation embedding of the main triple: 
\begin{equation}
	h_o=act(\bigoplus_{(s,r)\in\mathcal{N}(s)}(\{W_{\lambda(r)}\phi(\gamma(h_s,h_r), h_q) \})),
\end{equation} 
where $\gamma(\cdot)$ denote the weighted sum.
For the extra qualifier aggregator, 
QUAD introduces main triple embedding: \begin{equation}
	h_t=W(Cat(h_s,h_r,h_o))+b,
\end{equation}
and treat $h_t$ the same as entity embeddings to take part in the later message passing, enabling qualifiers to receive information from the main triple.

For the path-aggregated GNN-based KG embedding model (Path-GNN in \autoref{tab:metrics}) NBFNet \cite{zhu2021neural}, a distinct approach is employed. Rather than aggregating the neighborhood of a specific entity on the KG, NBFNet aggregates paths from one entity to another within its multi-hop graph neighborhood. The pair embedding is then created by aggregating path representations between entities $s$ and $o$ w.r.t. query relation $r$:
\vspace{-5px}
\begin{equation}
	h_r(s,o)=\bigoplus_{P\in\mathcal{P}(s,o)}(\phi(\prod_{i=1}^{|P|} h_{r_i})), 
\end{equation}
where $\mathcal{P}(s,o)$ denotes the set of path from entity $s$ to entity $o$; $h_{r_i}$ denotes the $i$-th relation within path $P$. 
The path-GNN is the most popular technique in current KGE research and has led to the development of many novel models \cite{zhang2023adaprop, chang2024path}.

Benefiting from the message passing mechanism \cite{kipf2022semi}, GNN-based models can directly receive graph information within the graph's multi-hop neighborhood.

In this paper, 
we focus on Cleaned JF17k \cite{galkin2020message} and WD50k \cite{galkin2020message}, 
two widely used datasets that have moderate qualifier density and graph scale, 
and are considered HKGs to have the most representative characteristics. 
Cleaned JF17k, which was proposed based on the original JF17k \cite{wen2016representation}, further addresses the data leakage issue.

On the other hand, we also conducted experiments with the Wikipeople \cite{guan2019link} and FBAUTO \cite{fatemi2021knowledge} datasets, which exhibit distinct HKG characteristics. As shown in \autoref{tab:dataset}, Wikipeople has a low qualifier density (less than 3\%), while FBAUTO is small in graph scale, containing only 2,094 unique entities and 8 unique relations.

\section{Model Performance on HKG} 
\label{sec:performance-hkg}
In this section, we experimented the performance of KGE and HKGE models on HKG, to see whether KGE can still perform silmilarly with HKGE nowadays. 
Different from most of previous researches, we data-wise decompose HKG into KG format, to analyse the capability of KGE models on HKG.
To achieve this, 
we propose the 3 decomposition methods in \autoref{sec:decomp}.
With the help of these methods, we carry out experiment with KGE models from 5 technical categories, interestingly, in \autoref{sec:findings}, we find some KGE can perform comparably with HKGE.
\subsection{Decomposition Methods}\label{sec:decomp}
Although \cite{alivanistos2021query} applied the DB reification method to convert HKG to KG format, as introduced, it can greatly shift the original HKG topology, even for the important main triples. 
As a result, we cannot analyse the HKG's each component with its DB reification method. 
Hence, to data-wise adapt KGE models to HKG and to better study the impact of different HKG information on KGE models, we design 3 simple decomposition methods that can preserve the HKG information to different extent:
\begin{itemize}[leftmargin=10pt]
\item \textbf{Prune method} prunes all qualifiers, only preserves the \textit{main-triple information}: 
\begin{equation}
	T_{prune}:=\{(s,r,o)\};
\end{equation}
\item \textbf{Direct method} directly links the qualifiers to the subject entity of the main triple: 
\begin{equation} 
T_{direct}:=\{(s,r,o),\{(s,qr_i,qe_i)\}_{i=1}^n\}, 
\end{equation}
compared with the prune method, it further considers qualifiers the same as main triples and further preserve the \textit{intra-qualifier information}, including qualifier entities and relations.
\item \textbf{Hyper method} creates new relation $r||qr_i$ for every qualfier $(qr_i,qe_i)$ with the name concatenation of relation $r$ and qualifier relation $qr_i$: 
\begin{align} T_{hyper}:=\{(s,r,o),\{(s,qr_i,qe_i)\}_{i=1}^n,\notag\\ \{(qe_i,r||qr_i,o)\}_{i=1}^n\}.
\end{align}
Compared to the direct method, this approach better preserves \textit{inter-qualifier information}, including the relation between qualifiers and the main triple, as well as between qualifiers within the same hyper-relational fact.
\end{itemize}
Overall, the decomposition methods keep the KGE models' original receptive field and make KGE models aggregate qualifiers similarly with aggregating main-triple information. 

\begin{table}[t]
	\centering
	\caption{HKG Dataset Statistics. 
		``\#Qual.'' denotes the range of the number of qualifier entity-relation pairs,
		``Tri.'' denotes the triplet facts with no qualifiers,
		and
		``HR'' denotes the hyper-relational facts.
	}
	\label{tab:dataset}
 \vspace{-10px}
	\setlength\tabcolsep{1.5pt}
	\resizebox{1.03\linewidth}{!}{
		\begin{tabular}{l|c|c|c|c|c|c|c|c} 
			\hline
			\bf Dataset   & $|\mathcal{V}|$ & $|\mathcal{R}|$ & \#Qual. & \#Tra.  & \#Val. & \#Tst. & \#Tri.  & \#HR    \\ 
			\hline
			Cleaned JF17k \cite{galkin2020message} & 25,092          & 320             & 0-4     & 49,120  & 12,280 & 17,635 & 54,551  & 24,484  \\
			WD50k \cite{galkin2020message}         & 47,155          & 531             & 0-65    & 166,435 & 23,913 & 46,159 & 204,340 & 32,167  \\
			Wikipeople \cite{guan2019link}    & 31,038          & 171             & 0-7     & 262,301 & 33,838 & 33,806 & 257,693 & 4,608   \\
			FBAUTO \cite{fatemi2021knowledge}        & 2,094           & 8               & 0-3     & 6,778   & 2,255  & 2,180  & 3,786   & 7,427   \\
			\hline
		\end{tabular}
	}
	\vspace{-10px}
\end{table}
\subsection{HKG datasets}

\subsection{Interesting Findings}\label{sec:findings}
In this paper, we evaluate several HKGE models alongside KGE models using our simple decomposition methods, as shown in \autoref{tab:metrics}. The results for NaLP\_fix and Hinge on WD50k are taken from \cite{xiong2023shrinking}, while the results for ComplEx, ConvE, and TransH were reproduced using PyKeen \cite{ali2021pykeen}. 
Other results were obtained from the original GitHub repositories. Since StarE, QUAD, and HyTransformer use the Transformer decoder, we replace the decoder in CompGCN with a Transformer to ensure consistency with these models.
All experiments were conducted on a single NVIDIA A100-40GB GPU, and parameters were either kept consistent with the original repositories or tuned using grid search when dataset-specific parameters were not available.

As shown in \autoref{tab:metrics}, compared to more advanced GNN-based KGE models, Tensor Decomposition (TD)-based KGE models are significantly outperformed by their respective HKGE extensions. 
On one hand, when applied to the pruned method, the HKG main triples differ from the KG main triples in that they do not contain complete information about HKG facts, as they lack supporting qualifier information. 
As a result, when coping TD-based KGE models with HKG decomposition methods, the main-triple embeddings were affected by the information that should exist in the qualifiers. Hence, the decomposition from the basis matrix into the main triple's embedding makes no sense and can severely bring noise.
On the other hand, the GNN-based KGE models aggregate the respective graph neighborhood for generating main triple embeddings. In that case, the qualifier information just missing, rather than being simply injected into the main triple embedding and serving as noise. Hence, the negative effect of the prune decomposition method is less severe. 
Furthermore, since GNN-based KGE models separate their encoder and decoder modules, 
their performance can benefit from HKG-oriented decoders \cite{alivanistos2021query} when using the direct and hyper methods. This makes GNN-based KGE models the most effective choice for applying to HKGs.

More importantly, when applying the three decomposition methods, the KGE models CompGCN \cite{vashishth2019composition} and NBFNet \cite{zhu2021neural} achieve performance that is comparable to, or even better than, some popular HKGE models, 
such as CompGCN's HKG extension StarE \cite{galkin2020message}. 
These results introduce a need to further analyze the strengths and weaknesses of some of the best-performed KGE and HKGE models, and raise two questions.
\begin{enumerate}[leftmargin=*]
\item Consider the key difference between KGE and HKGE is whether they apply specially designed modules to capture qualifiers, is it necessary to develop various HKGEs?
\item Why would current HKGE models not perform better? What important information are they overlooking?
\end{enumerate}
To answer these questions, we study current KG and HKG models regarding their qualifier capture and graph reception ability in the next section. 
\begin{table}[t]
	\centering
	\caption{HKG information preserved by different decomposition methods.}
	\label{tab:decomp}
	\vspace{-10px}
	\resizebox{\linewidth}{!}{%
		\begin{tabular}{l|c|c|c} 
			\toprule
			\multicolumn{1}{c|}{\multirow{2}{*}{\textbf{Method }}} & \multicolumn{3}{c}{\textbf{Preserved Information}}                         \\ 
			\cline{2-4}
			\multicolumn{1}{c|}{}          & \textbf{Main Triple} & \textbf{Intra-Qualifier} &  \textbf{Inter-Qualifier}\\ 
			\midrule
			Prune                                                  &            \checkmark          &                          &                       \\
			Direct                                                 &          \checkmark            &                     \checkmark     &                       \\
			Hyper                                           &            \checkmark         &           \checkmark              &           \checkmark           \\
			DB reification \cite{alivanistos2021query}                                                  &                  &           \checkmark              &           \checkmark           \\
			\bottomrule
		\end{tabular}
	}
	\vspace{-12px}
\end{table}

\begin{figure*}[t]
	\centering
	\includegraphics[width=1\linewidth]{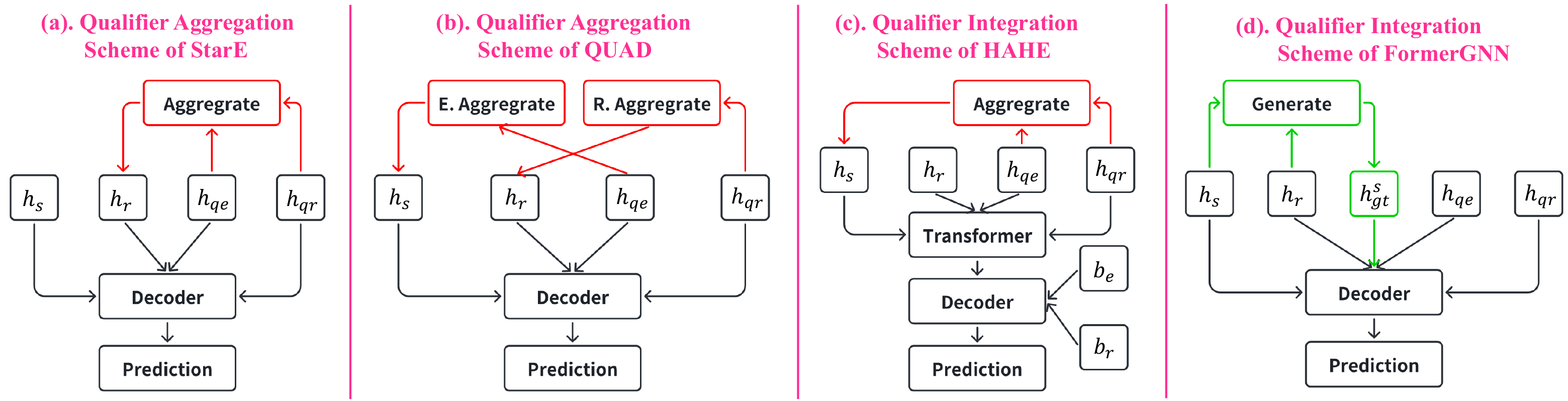}
	\vspace{-10px}
	\caption{
		The qualifier aggregation scheme of StarE, QUAD, and HAHE compared with our FormerGNN framework, where E. Aggregate and R. Aggregate represent QUAD's qualifier entity and relation aggregation operation, respectively. Compared with StarE, QUAD, and HAHE, FormerGNN does not aggregate qualifier information in $h_{qe}$ and $h_{qr}$ into the main-triple's embeddings $h_s$ or $h_r$. Instead, FormerGNN stores the graph's long-range dependency information separately in $h^s_{gt}$, then jointly and directly carries out prediction with $h^s_{gt}$, and information from the main-triple and qualifiers embedding matrixes.
	}
	\label{fig:QAggregation}
	\vspace{-10px}
\end{figure*}

\section{Qualifier Capture and Graph Reception}
\label{sec:futher}
As introduced, to answer two questions raised in the previous section, we need to analyze current KGE and HKGE models' ability from two parts: Qualifier Capture and Graph Reception. 
\begin{itemize}[leftmargin=10px]
\item In \autoref{sec:integration}, after analyzing the performance of several best-performed KGE models with the three decomposition methods, we claim it is still necessary to develop more novel HKGEs, as the ability of decomposition methods to preserve the HKG information is still insufficient.
\item In \autoref{sec:reception}, after analyzing the graph reception ability of current HKGE models, we find these models are either insufficient in capturing the graph's long-range dependency or suffer from information compression issues when aggregating qualifiers with the main triple.
\end{itemize}

\subsection{Qualifier Capture}\label{sec:integration}
In this section, we analyze the ability of KGE and HKGE models to capture qualifiers to address the question of whether it is necessary to develop various HKGEs. Based on the following observations, we argue that HKGEs are still needed, as the decomposition methods are insufficient in integrating qualifiers.
\begin{itemize}[leftmargin=10px]
\item Firstly, the path-GNN KGE model NBFNet, when preserving more HKG information, can suffer from a performance drop on Cleaned JF17k and WD50k.

\item Secondly, for neighbor-GNN KGE model CompGCN, when preserving more information, its performance remains similar or only marginally increase on two most representative dataset. 

\item Thirdly, the Transformer-based HKGE model HyperFormer achieves the SOTA performance 
on two most representative dataset, 
this also shows the importance of a properly and specially designed qualifier capturing mechanism, to learn the qualifier together with the main triple, rather than inject one into another. 
\end{itemize}

These phenomena occur because qualifiers are often contain less important information than the main triple \cite{rosso2020beyond}. Hence, the direct and hyper methods may introduce noise paths or neighborhoods formed by qualifiers, which can interfere with capturing the more important main-triple information.

To further support our claim, we enhance the qualifier capture approach of the top-performing KGE model, NBFNet, 
resulting in an improved model called FormerGNN. FormerGNN integrates a Transformer-based qualifier capturing module with NBFNet to capture the original HKG topology. A more detailed analysis of FormerGNN are in \autoref{sec:formerGNN}.
\begin{figure*}[!ht]
	\centering
	\includegraphics[width=1\linewidth]{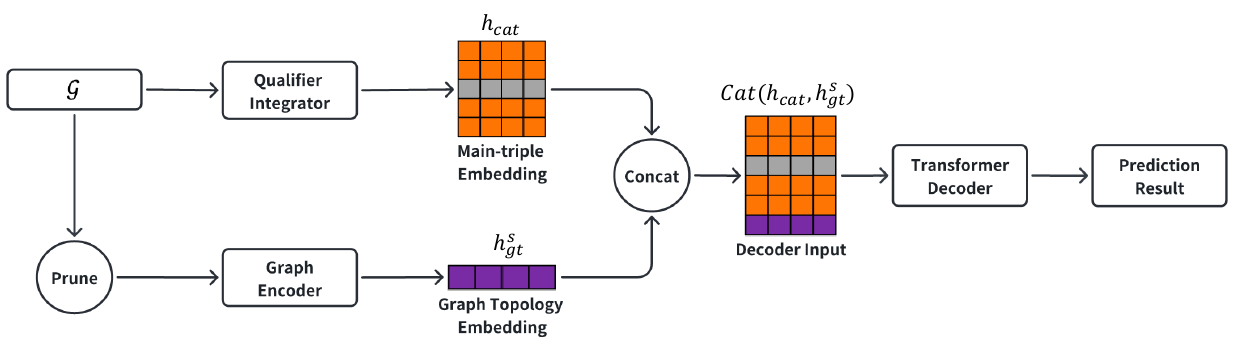}
	\vspace{-15px}
	\caption{
		The pipeline the FormerGNN framework, it involves concatenating the main-triple and qualifier embeddings, generated by the Qualifier Integrator with the original HKG topology, and the graph topology embeddings generated by the Graph Encoder with the HKG after applying the prune decomposition method.
	}
	\label{fig:formergnn}
	\vspace{-5px}
\end{figure*}
\begin{table*}[t]
\caption{The performance of FormerGNN and several best performed KG and HKG embedding models on 4 HKG datasets.}
\vspace{-8px}
\centering
\scalebox{0.9}{
\setlength{\tabcolsep}{3pt}{
	\resizebox{1\linewidth}{!}{%
		\begin{tabular}{l|ccc|ccc|ccc|ccc} 
			\hline\hline
			\multirow{2}{*}{\textbf{Model}} & \multicolumn{3}{c|}{\textbf{Cleaned JF17k~\cite{galkin2020message}}}      & \multicolumn{3}{c|}{\textbf{WD50k~\cite{galkin2020message}}}              & \multicolumn{3}{c|}{\textbf{Wikipeople~\cite{guan2019link}}}         & \multicolumn{3}{c}{\textbf{FBAUTO~\cite{fatemi2021knowledge}}}               \\ 
			\cline{2-13}
			& MRR            & H@1            & H@10           & MRR            & H@1            & H@10           & MRR            & H@1            & H@10           & MRR            & H@1            & H@10            \\ 
			\hline\hline
			StarE~\cite{galkin2020message}                           & 0.381          & 0.284          & 0.577          & 0.346          & 0.268          & 0.494          & 0.491          & 0.401          & 0.646          & 0.346          & 0.268          & 0.494           \\
			QUAD~\cite{shomer2023learning}                            & 0.402          & 0.305          & 0.602          & 0.349          & 0.275          & 0.487          & 0.493          & 0.422          & 0.619          & \textbf{0.868} & \textbf{0.836} & \textbf{0.921}  \\
			HAHE~\cite{luo2023hahe}                &   0.405                    & 0.310                    & 0.585                 & 0.360                    & 0.288                    & 0.496          &     0.479     &  \uline{0.430} &    0.627       &    0.848          & 0.817          & \uline{0.906}   \\
			HyperFormer \cite{hu2023hyperformer}                     & \uline{0.425}  & \uline{0.328}  & 0.613          & \uline{0.372}  & \uline{0.293}  & 0.522          & 0.488          & 0.382          & \uline{0.652}          & 0.819          & 0.789          & 0.873           \\
			NBFNet + Prune \cite{zhu2021neural}                 & 0.415          & 0.313          & \textbf{0.630} & 0.360          & 0.271          & \textbf{0.530} & \textbf{0.513} & \textbf{0.444} & 0.644  & 0.767          & 0.700          & 0.888           \\ 
			CompGCN + Hyper \cite{vashishth2019composition}                 & 0.389          & 0.294          & 0.590 & 0.350          & 0.277          & 0.488 & 0.511 & 0.439 & 0.639  & 0.857          & 0.834          & 0.903           \\ 
			\hline
			\bf FormerGNN                       & \textbf{0.432} & \textbf{0.335} & \uline{0.624}  & \textbf{0.377} & \textbf{0.299} & \uline{0.527}  & \uline{0.494}  & 0.392          & \textbf{0.653} & \uline{0.855}  & \uline{0.828}  & 0.903           \\
			\hline\hline
		\end{tabular}
		}}}
		\label{tab:formergnn}
		\vspace{-5px}
\end{table*}

\subsection{Graph Reception}\label{sec:reception}
To point out what important information are current HKGE models overlooking, in this section, we analyze current KGE and HKGE models' graph reception ability. 
With the following observations from \autoref{tab:metrics}, we claim some of the current HKGEs overlook the capture graph's long-range dependency.
\begin{itemize}[leftmargin=10px]
\item Firstly, with an additional GNN graph encoder capturing the graph's long-range dependency, QUAD \cite{shomer2023learning}, and even CompGCN \cite{vashishth2019composition} achieves better performance than HyTransformer \cite{yu2021improving} on Cleaned JF17k and WD50k,
which does not align with the claim of previous researches \cite{yu2021improving,hu2023hyperformer}.
\item Secondly, by employing the latest path-GNN KGE technique, NBFNet \cite{zhu2021neural} is the best at capturing the graph's long-range dependencies among all tested KGE models. 
As a result, it achieves the second-best performance on Cleaned JF17k and WD50k, even when using the prune method and lack of qualifier information.
\end{itemize}
We claim the reason behind these results as these models only achieve a limited receptive field. Specifically, a wide receptive field is beneficial for capturing the graph's long-range dependencies, which in turn contribute to improved model performance. Although the reason behind these results has been widely studied in the KGE field \cite{zhu2021neural, zhang2023adaprop}, on HKGs, researchers usually overlook the importance of a wide receptive field and consider the qualifiers as a substitution \cite{yu2021improving,hu2023hyperformer,xiong2023shrinking,wang2021link}. In this paper, we argue that a wide receptive field is also crucial for HKGs as a complement to qualifier information, rather than a substitution. As the main-triples typically contain much more important information than qualifiers \cite{rosso2020beyond}, the long-range dependencies formed by main-triples are also crucial.


On the other hand, although some HKGE models have wide receptive fields and better capture the graph's long-range dependencies, as shown in \autoref{fig:QAggregation}, they can suffer from information compression issues. This occurs when they aggregate qualifier information in $h_{qe}$ and $h_{qr}$ into the fixed-sized main-triple entity and relation embedding matrices 
$h_s$ and $h_r$. This not only limits the models' ability to integrate qualifiers but also introduces noise into the main-triple embedding matrices 
where the graph's long-range dependencies are stored. 
Our claim can be justified with the following observations from \autoref{tab:metrics}: 

\begin{itemize}[leftmargin=10px]
\item Firstly, although having a wide receptive field, the performance of StarE, QUAD and HAHE still falls behind HyperFormer on Cleaned JF17k and WD50k dataset. 
\item Secondly, the KGE model CompGCN with the hyper method can also perform comparably or even better than its HKG extension StarE and QUAD. 
\end{itemize}
The reason for these results is that StarE, QUAD, and HAHE simply aggregate qualifiers into the fixed-sized entity or relation embedding matrix. This aggregation compresses the noise from qualifiers into the main-triple embeddings. For StarE and QUAD, these embeddings were later used for GNN message passing, which even further affects the model’s ability to capture the information of the whole graph. 
This issue, known as over-squashing \cite{topping2021understanding} in KG (Further details of over-squashing are discussed in \autoref{sec:appendix_oversquashing}), can be even more pronounced in HKGs, as qualifier information can overwhelm the more important main triple information in the generated embedding matrices.

To further justify our claim, as well as propose a future research direction, we propose FormerGNN in \autoref{sec:formerGNN}, which can be considered a qualifier integrator plus a graph encoder to capture the graph's long-range dependency formed by main-triples, as well as the best-performed KGE NBFNet with a better approach to integrating qualifier and main-triple. 

\section{FormerGNN}\label{sec:formerGNN}


As discussed in previous sections, to support our claims and suggest a future direction for HKGE research, we need to capture the original HKG topology, achieve a wide receptive field, and improve the integration of qualifiers and main triples while avoiding information compression. To address these challenges, we propose the FormerGNN framework.
FormerGNN combines a Transformer-based qualifier integrator to capture qualifier information from the original HKG, along with a GNN-based graph encoder to capture the graph’s long-range dependencies. Finally, it performs predictions by jointly using main-triple and qualifier information without aggregating them into fix-sized embedding matrixes.
\begin{itemize}[leftmargin=10px]
\item Firstly, FormerGNN's graph encoder $GE$ would generate a graph topology embedding matrix $h^{s}_{gt}$ of the subject entity $s$ with the decomposed HKG by the prune method $T_{prune}$: 
\begin{equation}
h^{s}_{gt}=GE(T_{prune}(\mathcal{G})).
\end{equation}
Where $GE$ would generate the embedding of all main-triple entities, and then apply a linear transformation to make the generated matrix align with other embedding matrixes generated by the qualifier integrator.
\item Secondly, the graph topology embedding matrix would be concatenated with the main-triple embedding matrixes $h_s$, $h_r$, $h_{[MSK]}$ and qualifier embedding matrixes $h_{qr}$, $h_{qe}$, $\cdots$, from the qualifier integrator $QI$ to form the decoder input $h_{cat}$ for joint prediction: 
\begin{equation}
h_s,h_r,h_{[MSK]},h_{qr_1},h_{qe_1},\cdots \leftarrow QI(F)
\end{equation}
\begin{equation}
h_{cat}=Cat(h_s,h_r,h_{[MSK]},h_{qr_1},h_{qe_1},\cdots ,h^{s}_{gt}).
\end{equation}

\item Lastly, this concatenation would be passed to the transformer decoder to generate the distribution $D_\mathcal{V}$ over the graph node set $\mathcal{V}$ from the embedding matrix of the mask token $[MSK]$, avoid any direct aggregation: 
\begin{equation}
D_\mathcal{V}=Trm(h_{cat})^{[MSK]} \odot V.
\end{equation}
\end{itemize}

As shown in \autoref{tab:formergnn}, FormerGNN benefits from the advancements discussed earlier and shows improved performance compared to current HKGE models. When using NBFNet as the graph encoder and HyperFormer as the qualifier integrator, FormerGNN achieves state-of-the-art performance on both Cleaned JF17k and WD50k, highlighting the importance of the previously mentioned aspects. 
Note that both the qualifier integrator and the graph encoder can be substituted with more advanced transformer-based models and KG encoders. FormerGNN can easily benefit from the advances in fact or node level HKG and KGE research. 

On Wikipeople, which consists mostly of triples without qualifiers, the HKG-specific module of HyperFormer struggles to train effectively due to the lack of qualifiers, resulting in sub-optimal performance compared to KGE models NBFNet and CompGCN that apply appropriate decomposition methods. However, FormerGNN still achieves comparable or better performance than all other tested HKGE models.

On FBAUTO, which only has over 8000 entities and 8 relations, the lightweight QUAD model performs best. However, FormerGNN still outperforms both HyperFormer and NBFNet on this dataset. These results further emphasize the significance of capturing HKG's original topology, capturing long-range dependencies, and integrating qualifier information without simple aggregation.

\section{Conclusion}
In this paper, we investigate the performance of embedding models on HKGs to determine whether HKGEs notable performance is due to the KGE base model or the specially designed qualifier processing module. To conduct this study, we convert HKGs into KG format with three decomposition methods that preserve HKG information to varying degrees.

The results show that some novel GNN-based KGE models perform comparably to classic HKGE models. However, upon further analysis, we find that HKGE models are still necessary, as the decomposition methods disrupt the original HKG topology and can lead to information loss. Additionally, the sub-optimal performance of HKGE models can be attributed to either insufficient capture of the graph's long-range dependencies or the information compression issue caused by their aggregation of qualifier information into fix-sized main-triple embedding matrices.


To further support our claims and suggest a potential future direction for HKGE research, we propose the FormerGNN framework. This framework addresses the issues present in current KGE and HKGE models applied to HKGs. Experimental results show that FormerGNN achieves comparable or better performance than existing KGE and HKGE models on HKGs, and can easily benefit from the future advances of KGE models as the graph encoder.

\clearpage
\begin{appendix}
\section{Appendix: Over-squashing}\label{sec:appendix_oversquashing}

\begin{figure}[t]
	\centering
	\subfigure[\small{Cleaned JF17k}]{
		\label{fig:cur_jf17k}
		\includegraphics[width=0.45\linewidth]{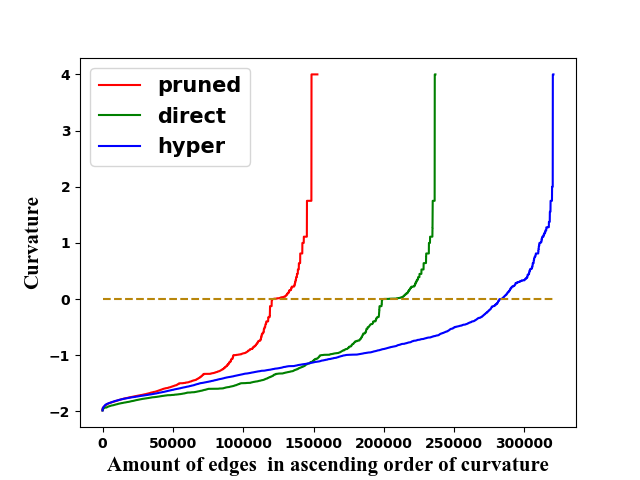}
	} \hspace{2mm}
	\subfigure[\small{Wikipeople}]{
		\label{fig:cur_wiki}
		\includegraphics[width=0.45\linewidth]{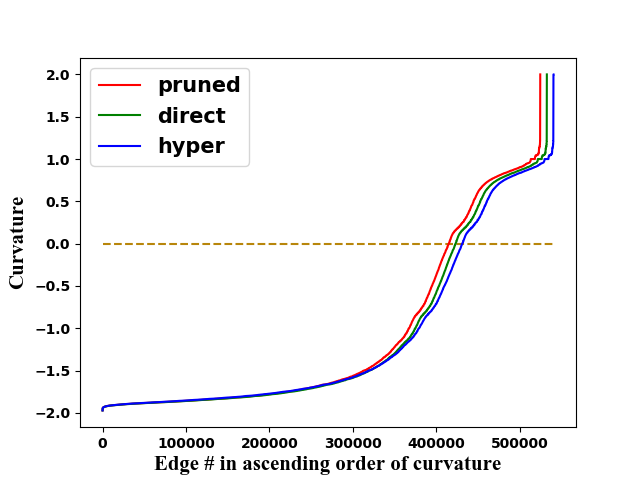}
	} \\ \vspace{-10px}
	\subfigure[\small{WD50k}]{
		\label{fig:cur_wd50k}
		\includegraphics[width=0.45\linewidth]{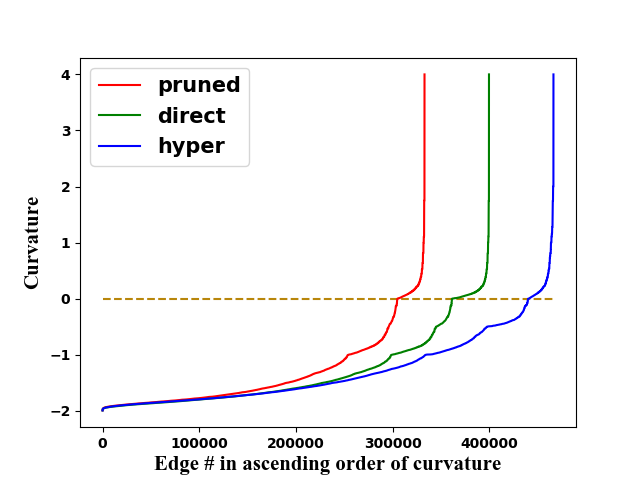}
	} \hspace{2mm}
	\subfigure[\small{FBAUTO}]{
		\label{fig:cur_fbauto}
		\includegraphics[width=0.45\linewidth]{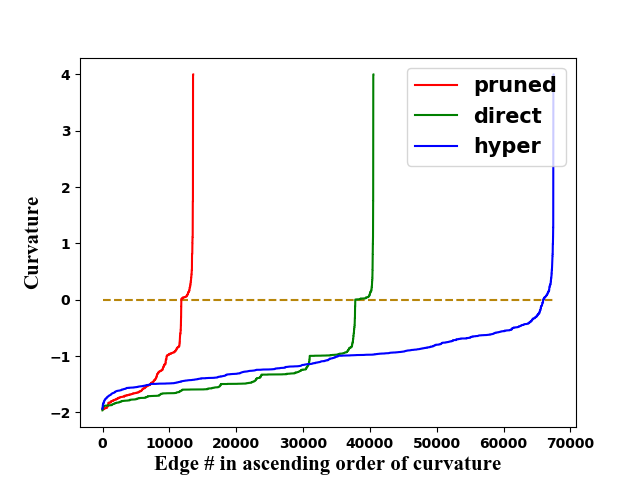}
	}
	\vspace{-10px}
	\caption{Curvature distribution for three decomposition methods on four datasets}
	\vspace{-15px}
	\label{fig:cur_4_dataset}
\end{figure}
In detail, the over-squashing issue refers to the phenomenon of severe information compression that occurs in neighbor-aggregated GNNs when propagating messages from distant entities, which can lead to a loss of graph's multi-hop topology information.
It tends to occur 
because
exponentially expanded amount of messages from distant 
entities
need to be compressed 
into fixed-sized vectors when passing to the target 
entity
\cite{alon2021bottleneck, topping2021understanding}. Over-squashing has become one of the main issues that hinder neighbor-GNNs' performance. 
Since our direct and hyper decomposition methods partially introduce new nodes and edges to the graph topology and treat these newly introduced triples in the same manner as main-triples, they can further exacerbate the over-squashing issue compared to other GNN-based HKG embedding methods, such as StarE and QUAD.

To better access the over-squashing issue in a quantitative way,
\cite{topping2021understanding}
conducts thorough analysis on graph topology and proposes the 
Balanced Forman curvature:
\begin{align}\label{ric}
	Ric(s,o)=&\frac{2}{d_s}+\frac{2}{d_o}-2+2\frac{|\#_\Delta(s,o)|}{max(d_s,d_o)}+\frac{|\#_\Delta(s,o)|}{min(d_s,d_o)} \\& \notag+\frac{\gamma_{max}^{-1}(s,o)}{max(d_s,d_o)}(|\#_\square^s|+|\#_\square^o|),
\end{align}
where 
$d_s$ is the degree of entity $s$,
$\#\Delta(s,o)$ are the triangles based on edge $(s,o)$, 
$\#\square^s$ are the neighbors of entity $o$ forming a 4-cycle based on $(s,o)$ without diagonals inside,
$|\cdot|$ is the set size,
$\gamma_{max}(s,o)$ is the maximal number of 4-cycles based on $(s,o)$ traversing a common entity, 
and 
$Ric(s,o) \in (-2, \infty)$,
Particularly, 
when
$Ric(s,o)<0$, 
it may
refer to the occurance of over-squashing on certain edge $(s,o)$.

\autoref{fig:cur_4_dataset} displays the curvature distribution of three decomposition methods across four datasets. 
The x-tick represents the edges numbered in ascending order of their curvature, the y-tick represents the curvature of edges.
When the curvature of an edge is less than or equal to 0, 
we consider it to have an over-squashing issue. 
Considering the prune decomposition as the baseline, the gap between other curves (the \textcolor{green}{green} and \textcolor{blue}{blue} curve) and its respective curve (the \textcolor{red}{red} curve) can be used to assess the level of exacerbation introduced by other decomposition methods when incorporating qualifiers in the same way as main-triples. Notably, the Cleaned JF17k and FBAUTO datasets exhibit the largest gaps, as they comprise the largest proportion of hyper-relational facts. In contrast, the Wikipeople dataset retains the smallest gap, as it contains only 2.6\% of hyper-relational facts.

\end{appendix}


\section*{GenAI Usage Disclosure}
The authors only used Grammarly to improve this paper's spelling, grammar, punctuation, clarity, and engagement. No sections or subsections in this paper are solely generated by AI.
\bibliographystyle{ACM-Reference-Format}
\bibliography{reference.bib}


\begin{thebibliography}{50}


\ifx \showCODEN    \undefined \def \showCODEN     #1{\unskip}     \fi
\ifx \showDOI      \undefined \def \showDOI       #1{#1}\fi
\ifx \showISBNx    \undefined \def \showISBNx     #1{\unskip}     \fi
\ifx \showISBNxiii \undefined \def \showISBNxiii  #1{\unskip}     \fi
\ifx \showISSN     \undefined \def \showISSN      #1{\unskip}     \fi
\ifx \showLCCN     \undefined \def \showLCCN      #1{\unskip}     \fi
\ifx \shownote     \undefined \def \shownote      #1{#1}          \fi
\ifx \showarticletitle \undefined \def \showarticletitle #1{#1}   \fi
\ifx \showURL      \undefined \def \showURL       {\relax}        \fi
\providecommand\bibfield[2]{#2}
\providecommand\bibinfo[2]{#2}
\providecommand\natexlab[1]{#1}
\providecommand\showeprint[2][]{arXiv:#2}

\bibitem[Ali et~al\mbox{.}(2021)]%
        {ali2021pykeen}
\bibfield{author}{\bibinfo{person}{Mehdi Ali}, \bibinfo{person}{Max
  Berrendorf}, \bibinfo{person}{Charles~Tapley Hoyt}, \bibinfo{person}{Laurent
  Vermue}, \bibinfo{person}{Sahand Sharifzadeh}, \bibinfo{person}{Volker
  Tresp}, {and} \bibinfo{person}{Jens Lehmann}.}
  \bibinfo{year}{2021}\natexlab{}.
\newblock \showarticletitle{{PyKEEN 1.0: A Python Library for Training and
  Evaluating Knowledge Graph Embeddings}}.
\newblock \bibinfo{journal}{\emph{Journal of Machine Learning Research}}
  \bibinfo{volume}{22}, \bibinfo{number}{82} (\bibinfo{year}{2021}),
  \bibinfo{pages}{1--6}.
\newblock
\urldef\tempurl%
\url{http://jmlr.org/papers/v22/20-825.html}
\showURL{%
\tempurl}


\bibitem[Alivanistos et~al\mbox{.}(2022)]%
        {alivanistos2021query}
\bibfield{author}{\bibinfo{person}{Dimitrios Alivanistos}, \bibinfo{person}{Max
  Berrendorf}, \bibinfo{person}{Michael Cochez}, {and} \bibinfo{person}{Mikhail
  Galkin}.} \bibinfo{year}{2022}\natexlab{}.
\newblock \showarticletitle{Query Embedding on Hyper-Relational Knowledge
  Graphs}. In \bibinfo{booktitle}{\emph{International Conference on Learning
  Representations}}.
\newblock


\bibitem[Alon and Yahav(2021)]%
        {alon2021bottleneck}
\bibfield{author}{\bibinfo{person}{Uri Alon} {and} \bibinfo{person}{Eran
  Yahav}.} \bibinfo{year}{2021}\natexlab{}.
\newblock \bibinfo{title}{On the Bottleneck of Graph Neural Networks and its
  Practical Implications}.
\newblock
\newblock
\showeprint[arxiv]{2006.05205}~[cs.LG]


\bibitem[Chang et~al\mbox{.}(2024)]%
        {chang2024path}
\bibfield{author}{\bibinfo{person}{Heng Chang}, \bibinfo{person}{Jiangnan Ye},
  \bibinfo{person}{Alejo Lopez-Avila}, \bibinfo{person}{Jinhua Du}, {and}
  \bibinfo{person}{Jia Li}.} \bibinfo{year}{2024}\natexlab{}.
\newblock \showarticletitle{Path-based explanation for knowledge graph
  completion}. In \bibinfo{booktitle}{\emph{Proceedings of the 30th ACM SIGKDD
  Conference on Knowledge Discovery and Data Mining}}.
  \bibinfo{pages}{231--242}.
\newblock


\bibitem[Dettmers et~al\mbox{.}(2018)]%
        {dettmers2018convolutional}
\bibfield{author}{\bibinfo{person}{Tim Dettmers}, \bibinfo{person}{Pasquale
  Minervini}, \bibinfo{person}{Pontus Stenetorp}, {and}
  \bibinfo{person}{Sebastian Riedel}.} \bibinfo{year}{2018}\natexlab{}.
\newblock \showarticletitle{Convolutional 2d knowledge graph embeddings}. In
  \bibinfo{booktitle}{\emph{Proceedings of the AAAI conference on artificial
  intelligence}}, Vol.~\bibinfo{volume}{32}.
\newblock


\bibitem[Ding et~al\mbox{.}(2024)]%
        {ding2024temporal}
\bibfield{author}{\bibinfo{person}{Zifeng Ding}, \bibinfo{person}{Jingcheng
  Wu}, \bibinfo{person}{Jingpei Wu}, \bibinfo{person}{Yan Xia},
  \bibinfo{person}{Bo Xiong}, {and} \bibinfo{person}{Volker Tresp}.}
  \bibinfo{year}{2024}\natexlab{}.
\newblock \showarticletitle{Temporal Fact Reasoning over Hyper-Relational
  Knowledge Graphs}. In \bibinfo{booktitle}{\emph{Findings of the Association
  for Computational Linguistics: EMNLP 2024}}. \bibinfo{pages}{355--373}.
\newblock


\bibitem[Fatemi et~al\mbox{.}(2021)]%
        {fatemi2021knowledge}
\bibfield{author}{\bibinfo{person}{Bahare Fatemi}, \bibinfo{person}{Perouz
  Taslakian}, \bibinfo{person}{David Vazquez}, {and} \bibinfo{person}{David
  Poole}.} \bibinfo{year}{2021}\natexlab{}.
\newblock \showarticletitle{Knowledge hypergraphs: prediction beyond binary
  relations}. In \bibinfo{booktitle}{\emph{Proceedings of the Twenty-Ninth
  International Conference on International Joint Conferences on Artificial
  Intelligence}}. \bibinfo{pages}{2191--2197}.
\newblock


\bibitem[Galkin et~al\mbox{.}(2020)]%
        {galkin2020message}
\bibfield{author}{\bibinfo{person}{Mikhail Galkin}, \bibinfo{person}{Priyansh
  Trivedi}, \bibinfo{person}{Gaurav Maheshwari}, \bibinfo{person}{Ricardo
  Usbeck}, {and} \bibinfo{person}{Jens Lehmann}.}
  \bibinfo{year}{2020}\natexlab{}.
\newblock \showarticletitle{Message Passing for Hyper-Relational Knowledge
  Graphs}. In \bibinfo{booktitle}{\emph{Proceedings of the 2020 Conference on
  Empirical Methods in Natural Language Processing (EMNLP)}}.
  \bibinfo{pages}{7346--7359}.
\newblock


\bibitem[Guan et~al\mbox{.}(2020)]%
        {guan2020neuinfer}
\bibfield{author}{\bibinfo{person}{Saiping Guan}, \bibinfo{person}{Xiaolong
  Jin}, \bibinfo{person}{Jiafeng Guo}, \bibinfo{person}{Yuanzhuo Wang}, {and}
  \bibinfo{person}{Xueqi Cheng}.} \bibinfo{year}{2020}\natexlab{}.
\newblock \showarticletitle{Neuinfer: Knowledge inference on n-ary facts}. In
  \bibinfo{booktitle}{\emph{Proceedings of the 58th annual meeting of the
  association for computational linguistics}}. \bibinfo{pages}{6141--6151}.
\newblock


\bibitem[Guan et~al\mbox{.}(2019)]%
        {guan2019link}
\bibfield{author}{\bibinfo{person}{Saiping Guan}, \bibinfo{person}{Xiaolong
  Jin}, \bibinfo{person}{Yuanzhuo Wang}, {and} \bibinfo{person}{Xueqi Cheng}.}
  \bibinfo{year}{2019}\natexlab{}.
\newblock \showarticletitle{Link prediction on n-ary relational data}. In
  \bibinfo{booktitle}{\emph{The world wide web conference}}.
  \bibinfo{pages}{583--593}.
\newblock


\bibitem[Guan et~al\mbox{.}(2024)]%
        {guan2024look}
\bibfield{author}{\bibinfo{person}{Saiping Guan}, \bibinfo{person}{Jiyao Wei},
  \bibinfo{person}{Xiaolong Jin}, \bibinfo{person}{Jiafeng Guo}, {and}
  \bibinfo{person}{Xueqi Cheng}.} \bibinfo{year}{2024}\natexlab{}.
\newblock \showarticletitle{Look Globally and Reason: Two-stage Path Reasoning
  over Sparse Knowledge Graphs}. In \bibinfo{booktitle}{\emph{Proceedings of
  the 33rd ACM International Conference on Information and Knowledge
  Management}}. \bibinfo{pages}{695--705}.
\newblock


\bibitem[Hamilton(1986)]%
        {Hamilton1986TheRF}
\bibfield{author}{\bibinfo{person}{Richard~S. Hamilton}.}
  \bibinfo{year}{1986}\natexlab{}.
\newblock \showarticletitle{The Ricci flow on surfaces}.
\newblock
\urldef\tempurl%
\url{https://api.semanticscholar.org/CorpusID:115319278}
\showURL{%
\tempurl}


\bibitem[Han et~al\mbox{.}(2020)]%
        {han2020explainable}
\bibfield{author}{\bibinfo{person}{Zhen Han}, \bibinfo{person}{Peng Chen},
  \bibinfo{person}{Yunpu Ma}, {and} \bibinfo{person}{Volker Tresp}.}
  \bibinfo{year}{2020}\natexlab{}.
\newblock \showarticletitle{Explainable subgraph reasoning for forecasting on
  temporal knowledge graphs}. In \bibinfo{booktitle}{\emph{International
  conference on learning representations}}.
\newblock


\bibitem[Hou et~al\mbox{.}(2023)]%
        {hou2023temporal}
\bibfield{author}{\bibinfo{person}{Zhongni Hou}, \bibinfo{person}{Xiaolong
  Jin}, \bibinfo{person}{Zixuan Li}, \bibinfo{person}{Long Bai},
  \bibinfo{person}{Saiping Guan}, \bibinfo{person}{Yutao Zeng},
  \bibinfo{person}{Jiafeng Guo}, {and} \bibinfo{person}{Xueqi Cheng}.}
  \bibinfo{year}{2023}\natexlab{}.
\newblock \showarticletitle{Temporal knowledge graph reasoning based on n-tuple
  modeling}. In \bibinfo{booktitle}{\emph{Findings of the Association for
  Computational Linguistics: EMNLP 2023}}. \bibinfo{pages}{1090--1100}.
\newblock


\bibitem[Hu et~al\mbox{.}(2023)]%
        {hu2023hyperformer}
\bibfield{author}{\bibinfo{person}{Zhiwei Hu}, \bibinfo{person}{V{\'\i}ctor
  Guti{\'e}rrez-Basulto}, \bibinfo{person}{Zhiliang Xiang}, \bibinfo{person}{Ru
  Li}, {and} \bibinfo{person}{Jeff~Z Pan}.} \bibinfo{year}{2023}\natexlab{}.
\newblock \showarticletitle{HyperFormer: Enhancing entity and relation
  interaction for hyper-relational knowledge graph completion}. In
  \bibinfo{booktitle}{\emph{Proceedings of the 32nd ACM International
  Conference on Information and Knowledge Management}}.
  \bibinfo{pages}{803--812}.
\newblock


\bibitem[Huang and Zeng(2024)]%
        {huang2024rd}
\bibfield{author}{\bibinfo{person}{Yubo Huang} {and} \bibinfo{person}{Guosun
  Zeng}.} \bibinfo{year}{2024}\natexlab{}.
\newblock \showarticletitle{RD-P: A Trustworthy Retrieval-Augmented Prompter
  with Knowledge Graphs for LLMs}. In \bibinfo{booktitle}{\emph{Proceedings of
  the 33rd ACM International Conference on Information and Knowledge
  Management}}. \bibinfo{pages}{942--952}.
\newblock


\bibitem[Kamdem~Teyou et~al\mbox{.}(2024)]%
        {kamdem2024embedding}
\bibfield{author}{\bibinfo{person}{Louis~Mozart Kamdem~Teyou},
  \bibinfo{person}{Caglar Demir}, {and} \bibinfo{person}{Axel-Cyrille
  Ngonga~Ngomo}.} \bibinfo{year}{2024}\natexlab{}.
\newblock \showarticletitle{Embedding Knowledge Graphs in Function Spaces}. In
  \bibinfo{booktitle}{\emph{Proceedings of the 33rd ACM International
  Conference on Information and Knowledge Management}}.
  \bibinfo{pages}{1070--1079}.
\newblock


\bibitem[Kazemi and Poole(2018)]%
        {kazemi2018simple}
\bibfield{author}{\bibinfo{person}{Seyed~Mehran Kazemi} {and}
  \bibinfo{person}{David Poole}.} \bibinfo{year}{2018}\natexlab{}.
\newblock \showarticletitle{Simple embedding for link prediction in knowledge
  graphs}.
\newblock \bibinfo{journal}{\emph{Advances in neural information processing
  systems}}  \bibinfo{volume}{31} (\bibinfo{year}{2018}).
\newblock


\bibitem[Kipf and Welling(2022)]%
        {kipf2022semi}
\bibfield{author}{\bibinfo{person}{Thomas~N Kipf} {and} \bibinfo{person}{Max
  Welling}.} \bibinfo{year}{2022}\natexlab{}.
\newblock \showarticletitle{Semi-Supervised Classification with Graph
  Convolutional Networks}. In \bibinfo{booktitle}{\emph{International
  Conference on Learning Representations}}.
\newblock


\bibitem[Lehmann et~al\mbox{.}(2015)]%
        {lehmann2015dbpedia}
\bibfield{author}{\bibinfo{person}{Jens Lehmann}, \bibinfo{person}{Robert
  Isele}, \bibinfo{person}{Max Jakob}, \bibinfo{person}{Anja Jentzsch},
  \bibinfo{person}{Dimitris Kontokostas}, \bibinfo{person}{Pablo~N Mendes},
  \bibinfo{person}{Sebastian Hellmann}, \bibinfo{person}{Mohamed Morsey},
  \bibinfo{person}{Patrick Van~Kleef}, \bibinfo{person}{S{\"o}ren Auer},
  {et~al\mbox{.}}} \bibinfo{year}{2015}\natexlab{}.
\newblock \showarticletitle{Dbpedia--a large-scale, multilingual knowledge base
  extracted from wikipedia}.
\newblock \bibinfo{journal}{\emph{Semantic web}} \bibinfo{volume}{6},
  \bibinfo{number}{2} (\bibinfo{year}{2015}), \bibinfo{pages}{167--195}.
\newblock


\bibitem[Li et~al\mbox{.}(2025)]%
        {li2025hyper}
\bibfield{author}{\bibinfo{person}{Jiecheng Li}, \bibinfo{person}{Xudong Luo},
  \bibinfo{person}{Guangquan Lu}, {and} \bibinfo{person}{Shichao Zhang}.}
  \bibinfo{year}{2025}\natexlab{}.
\newblock \showarticletitle{Hyper-Relational Knowledge Representation Learning
  with Multi-Hypergraph Disentanglement}. In
  \bibinfo{booktitle}{\emph{Proceedings of the ACM on Web Conference 2025}}.
  \bibinfo{pages}{3288--3299}.
\newblock


\bibitem[Li et~al\mbox{.}(2024a)]%
        {li2024integrating}
\bibfield{author}{\bibinfo{person}{Lijie Li}, \bibinfo{person}{Hui Wang},
  \bibinfo{person}{Jiahang Li}, \bibinfo{person}{Xiaodi Xu},
  \bibinfo{person}{Ye Wang}, {and} \bibinfo{person}{Tao Ren}.}
  \bibinfo{year}{2024}\natexlab{a}.
\newblock \showarticletitle{Integrating Structure and Text for Enhancing
  Hyper-relational Knowledge Graph Representation via Structure Soft Prompt
  Tuning}. In \bibinfo{booktitle}{\emph{Proceedings of the 33rd ACM
  International Conference on Information and Knowledge Management}}.
  \bibinfo{pages}{1226--1234}.
\newblock


\bibitem[Li et~al\mbox{.}(2024b)]%
        {li2024learning}
\bibfield{author}{\bibinfo{person}{Zhuofeng Li}, \bibinfo{person}{Haoxiang
  Zhang}, \bibinfo{person}{Qiannan Zhang}, \bibinfo{person}{Ziyi Kou}, {and}
  \bibinfo{person}{Shichao Pei}.} \bibinfo{year}{2024}\natexlab{b}.
\newblock \showarticletitle{Learning from novel knowledge: Continual few-shot
  knowledge graph completion}. In \bibinfo{booktitle}{\emph{Proceedings of the
  33rd ACM International Conference on Information and Knowledge Management}}.
  \bibinfo{pages}{1326--1335}.
\newblock


\bibitem[Liu et~al\mbox{.}(2023)]%
        {liu2023self}
\bibfield{author}{\bibinfo{person}{Yi Liu}, \bibinfo{person}{Hongrui Xuan},
  \bibinfo{person}{Bohan Li}, \bibinfo{person}{Meng Wang},
  \bibinfo{person}{Tong Chen}, {and} \bibinfo{person}{Hongzhi Yin}.}
  \bibinfo{year}{2023}\natexlab{}.
\newblock \showarticletitle{Self-supervised dynamic hypergraph recommendation
  based on hyper-relational knowledge graph}. In
  \bibinfo{booktitle}{\emph{Proceedings of the 32nd ACM International
  Conference on Information and Knowledge Management}}.
  \bibinfo{pages}{1617--1626}.
\newblock


\bibitem[Liu et~al\mbox{.}(2021)]%
        {liu2021role}
\bibfield{author}{\bibinfo{person}{Yu Liu}, \bibinfo{person}{Quanming Yao},
  {and} \bibinfo{person}{Yong Li}.} \bibinfo{year}{2021}\natexlab{}.
\newblock \showarticletitle{Role-aware modeling for n-ary relational knowledge
  bases}. In \bibinfo{booktitle}{\emph{Proceedings of the Web Conference
  2021}}. \bibinfo{pages}{2660--2671}.
\newblock


\bibitem[Luo et~al\mbox{.}(2023)]%
        {luo2023hahe}
\bibfield{author}{\bibinfo{person}{Haoran Luo}, \bibinfo{person}{Yuhao Yang},
  \bibinfo{person}{Yikai Guo}, \bibinfo{person}{Mingzhi Sun},
  \bibinfo{person}{Tianyu Yao}, \bibinfo{person}{Zichen Tang},
  \bibinfo{person}{Kaiyang Wan}, \bibinfo{person}{Meina Song},
  \bibinfo{person}{Wei Lin}, {et~al\mbox{.}}} \bibinfo{year}{2023}\natexlab{}.
\newblock \showarticletitle{HAHE: Hierarchical Attention for Hyper-Relational
  Knowledge Graphs in Global and Local Level}.
\newblock \bibinfo{journal}{\emph{arXiv preprint arXiv:2305.06588}}
  (\bibinfo{year}{2023}).
\newblock


\bibitem[Pahuja et~al\mbox{.}(2024)]%
        {pahuja2024reviving}
\bibfield{author}{\bibinfo{person}{Vardaan Pahuja}, \bibinfo{person}{Weidi
  Luo}, \bibinfo{person}{Yu Gu}, \bibinfo{person}{Cheng-Hao Tu},
  \bibinfo{person}{Hong-You Chen}, \bibinfo{person}{Tanya Berger-Wolf},
  \bibinfo{person}{Charles Stewart}, \bibinfo{person}{Song Gao},
  \bibinfo{person}{Wei-Lun Chao}, {and} \bibinfo{person}{Yu Su}.}
  \bibinfo{year}{2024}\natexlab{}.
\newblock \showarticletitle{Reviving the Context: Camera Trap Species
  Classification as Link Prediction on Multimodal Knowledge Graphs}. In
  \bibinfo{booktitle}{\emph{Proceedings of the 33rd ACM International
  Conference on Information and Knowledge Management}}.
  \bibinfo{pages}{1825--1835}.
\newblock


\bibitem[Rosso et~al\mbox{.}(2020)]%
        {rosso2020beyond}
\bibfield{author}{\bibinfo{person}{Paolo Rosso}, \bibinfo{person}{Dingqi Yang},
  {and} \bibinfo{person}{Philippe Cudr{\'e}-Mauroux}.}
  \bibinfo{year}{2020}\natexlab{}.
\newblock \showarticletitle{Beyond triplets: hyper-relational knowledge graph
  embedding for link prediction}. In \bibinfo{booktitle}{\emph{Proceedings of
  the web conference 2020}}. \bibinfo{pages}{1885--1896}.
\newblock


\bibitem[Shehzad et~al\mbox{.}(2024)]%
        {shehzad2024graph}
\bibfield{author}{\bibinfo{person}{Ahsan Shehzad}, \bibinfo{person}{Feng Xia},
  \bibinfo{person}{Shagufta Abid}, \bibinfo{person}{Ciyuan Peng},
  \bibinfo{person}{Shuo Yu}, \bibinfo{person}{Dongyu Zhang}, {and}
  \bibinfo{person}{Karin Verspoor}.} \bibinfo{year}{2024}\natexlab{}.
\newblock \showarticletitle{Graph transformers: A survey}.
\newblock \bibinfo{journal}{\emph{arXiv preprint arXiv:2407.09777}}
  (\bibinfo{year}{2024}).
\newblock


\bibitem[Shomer et~al\mbox{.}(2023)]%
        {shomer2023learning}
\bibfield{author}{\bibinfo{person}{Harry Shomer}, \bibinfo{person}{Wei Jin},
  \bibinfo{person}{Juanhui Li}, \bibinfo{person}{Yao Ma}, {and}
  \bibinfo{person}{Hui Liu}.} \bibinfo{year}{2023}\natexlab{}.
\newblock \showarticletitle{Learning representations for hyper-relational
  knowledge graphs}. In \bibinfo{booktitle}{\emph{Proceedings of the
  International Conference on Advances in Social Networks Analysis and
  Mining}}. \bibinfo{pages}{253--257}.
\newblock


\bibitem[Topping et~al\mbox{.}(2022)]%
        {topping2021understanding}
\bibfield{author}{\bibinfo{person}{Jake Topping}, \bibinfo{person}{Francesco
  Di~Giovanni}, \bibinfo{person}{Benjamin~Paul Chamberlain},
  \bibinfo{person}{Xiaowen Dong}, {and} \bibinfo{person}{Michael~M Bronstein}.}
  \bibinfo{year}{2022}\natexlab{}.
\newblock \showarticletitle{Understanding over-squashing and bottlenecks on
  graphs via curvature}. In \bibinfo{booktitle}{\emph{International Conference
  on Learning Representations}}.
\newblock


\bibitem[Trouillon et~al\mbox{.}(2016)]%
        {trouillon2016complex}
\bibfield{author}{\bibinfo{person}{Th{\'e}o Trouillon},
  \bibinfo{person}{Johannes Welbl}, \bibinfo{person}{Sebastian Riedel},
  \bibinfo{person}{{\'E}ric Gaussier}, {and} \bibinfo{person}{Guillaume
  Bouchard}.} \bibinfo{year}{2016}\natexlab{}.
\newblock \showarticletitle{Complex embeddings for simple link prediction}. In
  \bibinfo{booktitle}{\emph{International conference on machine learning}}.
  PMLR, \bibinfo{pages}{2071--2080}.
\newblock


\bibitem[Vashishth et~al\mbox{.}(2019)]%
        {vashishth2019composition}
\bibfield{author}{\bibinfo{person}{Shikhar Vashishth}, \bibinfo{person}{Soumya
  Sanyal}, \bibinfo{person}{Vikram Nitin}, {and} \bibinfo{person}{Partha
  Talukdar}.} \bibinfo{year}{2019}\natexlab{}.
\newblock \showarticletitle{Composition-based Multi-Relational Graph
  Convolutional Networks}. In \bibinfo{booktitle}{\emph{International
  Conference on Learning Representations}}.
\newblock


\bibitem[Vaswani(2017)]%
        {vaswani2017attention}
\bibfield{author}{\bibinfo{person}{A Vaswani}.}
  \bibinfo{year}{2017}\natexlab{}.
\newblock \showarticletitle{Attention is all you need}.
\newblock \bibinfo{journal}{\emph{Advances in Neural Information Processing
  Systems}} (\bibinfo{year}{2017}).
\newblock


\bibitem[Vrande{\v{c}}i{\'c} and Kr{\"o}tzsch(2014)]%
        {vrandevcic2014wikidata}
\bibfield{author}{\bibinfo{person}{Denny Vrande{\v{c}}i{\'c}} {and}
  \bibinfo{person}{Markus Kr{\"o}tzsch}.} \bibinfo{year}{2014}\natexlab{}.
\newblock \showarticletitle{Wikidata: a free collaborative knowledgebase}.
\newblock \bibinfo{journal}{\emph{Commun. ACM}} \bibinfo{volume}{57},
  \bibinfo{number}{10} (\bibinfo{year}{2014}), \bibinfo{pages}{78--85}.
\newblock


\bibitem[Wang et~al\mbox{.}(2021)]%
        {wang2021link}
\bibfield{author}{\bibinfo{person}{Quan Wang}, \bibinfo{person}{Haifeng Wang},
  \bibinfo{person}{Yajuan Lyu}, {and} \bibinfo{person}{Yong Zhu}.}
  \bibinfo{year}{2021}\natexlab{}.
\newblock \showarticletitle{Link Prediction on N-ary Relational Facts: A
  Graph-based Approach}. In \bibinfo{booktitle}{\emph{Findings of the
  Association for Computational Linguistics: ACL-IJCNLP 2021}}.
  \bibinfo{pages}{396--407}.
\newblock


\bibitem[Wang et~al\mbox{.}(2014)]%
        {wang2014knowledge}
\bibfield{author}{\bibinfo{person}{Zhen Wang}, \bibinfo{person}{Jianwen Zhang},
  \bibinfo{person}{Jianlin Feng}, {and} \bibinfo{person}{Zheng Chen}.}
  \bibinfo{year}{2014}\natexlab{}.
\newblock \showarticletitle{Knowledge graph embedding by translating on
  hyperplanes}. In \bibinfo{booktitle}{\emph{Proceedings of the AAAI conference
  on artificial intelligence}}, Vol.~\bibinfo{volume}{28}.
\newblock


\bibitem[Wen et~al\mbox{.}(2016)]%
        {wen2016representation}
\bibfield{author}{\bibinfo{person}{Jianfeng Wen}, \bibinfo{person}{Jianxin Li},
  \bibinfo{person}{Yongyi Mao}, \bibinfo{person}{Shini Chen}, {and}
  \bibinfo{person}{Richong Zhang}.} \bibinfo{year}{2016}\natexlab{}.
\newblock \showarticletitle{On the representation and embedding of knowledge
  bases beyond binary relations}. In \bibinfo{booktitle}{\emph{Proceedings of
  the Twenty-Fifth International Joint Conference on Artificial Intelligence}}.
  \bibinfo{pages}{1300--1307}.
\newblock


\bibitem[Xiong et~al\mbox{.}(2023)]%
        {xiong2023shrinking}
\bibfield{author}{\bibinfo{person}{Bo Xiong}, \bibinfo{person}{Mojtaba
  Nayyeri}, \bibinfo{person}{Shirui Pan}, {and} \bibinfo{person}{Steffen
  Staab}.} \bibinfo{year}{2023}\natexlab{}.
\newblock \showarticletitle{Shrinking Embeddings for Hyper-Relational Knowledge
  Graphs}. In \bibinfo{booktitle}{\emph{Proceedings of the 61st Annual Meeting
  of the Association for Computational Linguistics (Volume 1: Long Papers)}}.
  \bibinfo{pages}{13306--13320}.
\newblock


\bibitem[Xiong et~al\mbox{.}(2017)]%
        {xiong2017explicit}
\bibfield{author}{\bibinfo{person}{Chenyan Xiong}, \bibinfo{person}{Russell
  Power}, {and} \bibinfo{person}{Jamie Callan}.}
  \bibinfo{year}{2017}\natexlab{}.
\newblock \showarticletitle{Explicit semantic ranking for academic search via
  knowledge graph embedding}. In \bibinfo{booktitle}{\emph{Proceedings of the
  26th international conference on world wide web}}.
  \bibinfo{pages}{1271--1279}.
\newblock


\bibitem[Yih et~al\mbox{.}(2015)]%
        {yih2015semantic}
\bibfield{author}{\bibinfo{person}{Scott Wen-tau Yih},
  \bibinfo{person}{Ming-Wei Chang}, \bibinfo{person}{Xiaodong He}, {and}
  \bibinfo{person}{Jianfeng Gao}.} \bibinfo{year}{2015}\natexlab{}.
\newblock \showarticletitle{Semantic parsing via staged query graph generation:
  Question answering with knowledge base}. In
  \bibinfo{booktitle}{\emph{Proceedings of the Joint Conference of the 53rd
  Annual Meeting of the ACL and the 7th International Joint Conference on
  Natural Language Processing of the AFNLP}}.
\newblock


\bibitem[Yu and Yang(2021)]%
        {yu2021improving}
\bibfield{author}{\bibinfo{person}{Donghan Yu} {and} \bibinfo{person}{Yiming
  Yang}.} \bibinfo{year}{2021}\natexlab{}.
\newblock \showarticletitle{Improving hyper-relational knowledge graph
  completion}.
\newblock \bibinfo{journal}{\emph{arXiv preprint arXiv:2104.08167}}
  (\bibinfo{year}{2021}).
\newblock


\bibitem[Yuan et~al\mbox{.}(2025)]%
        {yuan2025survey}
\bibfield{author}{\bibinfo{person}{Chaohao Yuan}, \bibinfo{person}{Kangfei
  Zhao}, \bibinfo{person}{Ercan~Engin Kuruoglu}, \bibinfo{person}{Liang Wang},
  \bibinfo{person}{Tingyang Xu}, \bibinfo{person}{Wenbing Huang},
  \bibinfo{person}{Deli Zhao}, \bibinfo{person}{Hong Cheng}, {and}
  \bibinfo{person}{Yu Rong}.} \bibinfo{year}{2025}\natexlab{}.
\newblock \showarticletitle{A survey of graph transformers: Architectures,
  theories and applications}.
\newblock \bibinfo{journal}{\emph{arXiv preprint arXiv:2502.16533}}
  (\bibinfo{year}{2025}).
\newblock


\bibitem[Yun et~al\mbox{.}(2019)]%
        {yun2019graph}
\bibfield{author}{\bibinfo{person}{Seongjun Yun}, \bibinfo{person}{Minbyul
  Jeong}, \bibinfo{person}{Raehyun Kim}, \bibinfo{person}{Jaewoo Kang}, {and}
  \bibinfo{person}{Hyunwoo~J Kim}.} \bibinfo{year}{2019}\natexlab{}.
\newblock \showarticletitle{Graph transformer networks}.
\newblock \bibinfo{journal}{\emph{Advances in neural information processing
  systems}}  \bibinfo{volume}{32} (\bibinfo{year}{2019}).
\newblock


\bibitem[Zhang et~al\mbox{.}(2016)]%
        {zhang2016collaborative}
\bibfield{author}{\bibinfo{person}{Fuzheng Zhang},
  \bibinfo{person}{Nicholas~Jing Yuan}, \bibinfo{person}{Defu Lian},
  \bibinfo{person}{Xing Xie}, {and} \bibinfo{person}{Wei-Ying Ma}.}
  \bibinfo{year}{2016}\natexlab{}.
\newblock \showarticletitle{Collaborative knowledge base embedding for
  recommender systems}. In \bibinfo{booktitle}{\emph{Proceedings of the 22nd
  ACM SIGKDD international conference on knowledge discovery and data mining}}.
  \bibinfo{pages}{353--362}.
\newblock


\bibitem[Zhang et~al\mbox{.}(2018)]%
        {zhang2018scalable}
\bibfield{author}{\bibinfo{person}{Richong Zhang}, \bibinfo{person}{Junpeng
  Li}, \bibinfo{person}{Jiajie Mei}, {and} \bibinfo{person}{Yongyi Mao}.}
  \bibinfo{year}{2018}\natexlab{}.
\newblock \showarticletitle{Scalable instance reconstruction in knowledge bases
  via relatedness affiliated embedding}. In
  \bibinfo{booktitle}{\emph{Proceedings of the 2018 world wide web
  conference}}. \bibinfo{pages}{1185--1194}.
\newblock


\bibitem[Zhang et~al\mbox{.}(2024b)]%
        {zhang2024language}
\bibfield{author}{\bibinfo{person}{Tianli Zhang}, \bibinfo{person}{Tongya
  Zheng}, \bibinfo{person}{Zhenbang Xiao}, \bibinfo{person}{Zulong Chen},
  \bibinfo{person}{Liangyue Li}, \bibinfo{person}{Zunlei Feng},
  \bibinfo{person}{Dongxiang Zhang}, {and} \bibinfo{person}{Mingli Song}.}
  \bibinfo{year}{2024}\natexlab{b}.
\newblock \showarticletitle{Language Models-enhanced Semantic Topology
  Representation Learning For Temporal Knowledge Graph Extrapolation}. In
  \bibinfo{booktitle}{\emph{Proceedings of the 33rd ACM International
  Conference on Information and Knowledge Management}}.
  \bibinfo{pages}{3227--3236}.
\newblock


\bibitem[Zhang et~al\mbox{.}(2023)]%
        {zhang2023adaprop}
\bibfield{author}{\bibinfo{person}{Yongqi Zhang}, \bibinfo{person}{Zhanke
  Zhou}, \bibinfo{person}{Quanming Yao}, \bibinfo{person}{Xiaowen Chu}, {and}
  \bibinfo{person}{Bo Han}.} \bibinfo{year}{2023}\natexlab{}.
\newblock \showarticletitle{Adaprop: Learning adaptive propagation for graph
  neural network based knowledge graph reasoning}. In
  \bibinfo{booktitle}{\emph{Proceedings of the 29th ACM SIGKDD Conference on
  Knowledge Discovery and Data Mining}}. \bibinfo{pages}{3446--3457}.
\newblock


\bibitem[Zhang et~al\mbox{.}(2024a)]%
        {zhang2024gail}
\bibfield{author}{\bibinfo{person}{Zhiqiang Zhang}, \bibinfo{person}{Liqiang
  Wen}, {and} \bibinfo{person}{Wen Zhao}.} \bibinfo{year}{2024}\natexlab{a}.
\newblock \showarticletitle{A gail fine-tuned llm enhanced framework for
  low-resource knowledge graph question answering}. In
  \bibinfo{booktitle}{\emph{Proceedings of the 33rd ACM International
  Conference on Information and Knowledge Management}}.
  \bibinfo{pages}{3300--3309}.
\newblock


\bibitem[Zhu et~al\mbox{.}(2021)]%
        {zhu2021neural}
\bibfield{author}{\bibinfo{person}{Zhaocheng Zhu}, \bibinfo{person}{Zuobai
  Zhang}, \bibinfo{person}{Louis-Pascal Xhonneux}, {and} \bibinfo{person}{Jian
  Tang}.} \bibinfo{year}{2021}\natexlab{}.
\newblock \showarticletitle{Neural bellman-ford networks: A general graph
  neural network framework for link prediction}.
\newblock \bibinfo{journal}{\emph{Advances in Neural Information Processing
  Systems}}  \bibinfo{volume}{34} (\bibinfo{year}{2021}),
  \bibinfo{pages}{29476--29490}.
\newblock


\end{thebibliography}
\end{document}